\documentclass[10pt,twocolumn,letterpaper]{article}

\usepackage{cvpr}
\usepackage{times}
\usepackage{epsfig}
\usepackage{graphicx}
\usepackage{amsmath}
\usepackage{amssymb}
\usepackage{enumitem}
\usepackage{bm}

\usepackage[breaklinks=true,bookmarks=false,colorlinks=true]{hyperref}

\cvprfinalcopy 

\def\cvprPaperID{****} 

\begin{document}

\title{Auto-Encoding Scene Graphs for Image Captioning}

\author{Xu Yang, Kaihua Tang, Hanwang Zhang, Jianfei Cai\\
School of Computer Science and Engineering,\\
Nanyang Technological University,\\
\tt\small{ \{s170018@e,kaihua001@e,hanwangzhang@,ASJFCai@\}.ntu.edu.sg}
}

\maketitle

\begin{abstract}
We propose Scene Graph Auto-Encoder (SGAE) that incorporates the language inductive bias into the encoder-decoder image captioning framework for more human-like captions. Intuitively, we humans use the inductive bias to compose collocations and contextual inference in discourse. For example, when we see the relation ``person on bike'', it is natural to replace ``on'' with ``ride'' and infer ``person riding bike on a road'' even the ``road'' is not evident. Therefore, exploiting such bias as a language prior is expected to help the conventional encoder-decoder models less likely overfit to the dataset bias and focus on reasoning. Specifically, we use the scene graph --- a directed graph ($\mathcal{G}$) where an object node is connected by adjective nodes and relationship nodes --- to represent the complex structural layout of both image ($\mathcal{I}$) and sentence ($\mathcal{S}$). In the textual domain, we use SGAE to learn a dictionary ($\mathcal{D}$) that helps to reconstruct sentences in the $\mathcal{S}\rightarrow \mathcal{G} \rightarrow \mathcal{D} \rightarrow \mathcal{S}$ pipeline, where $\mathcal{D}$ encodes the desired language prior; in the vision-language domain, we use the shared $\mathcal{D}$ to guide the encoder-decoder in the $\mathcal{I}\rightarrow \mathcal{G}\rightarrow \mathcal{D} \rightarrow \mathcal{S}$ pipeline. Thanks to the scene graph representation and shared dictionary, the inductive bias is transferred across domains in principle. We validate the effectiveness of SGAE on the challenging MS-COCO image captioning benchmark, \eg, our SGAE-based single-model achieves a new state-of-the-art $127.8$ CIDEr-D on the Karpathy split, and a competitive $125.5$ CIDEr-D (c40) on the official server even compared to other ensemble models.
\end{abstract}

\section{Introduction}
\begin{figure}[t]
\centering
\includegraphics[width=1\linewidth,trim = 5mm 5mm 5mm 5mm,clip]{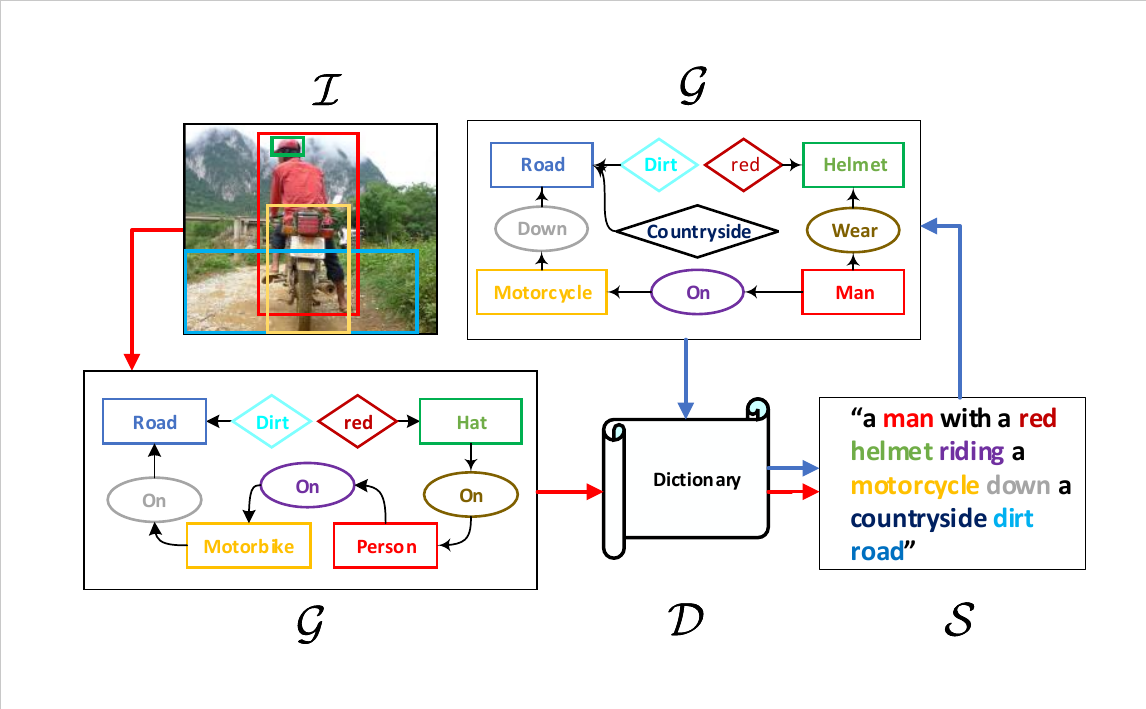}
   \caption{Illustration of auto-encoding scene graphs (blue arrows) into the conventional encoder-decoder framework for image captioning (red arrows), where the language inductive bias is encoded in the trainable shared dictionary. Word colors correspond to nodes in image and sentence scene graphs.}
\label{fig:1}
\end{figure}
Modern image captioning models employ an end-to-end encoder-decoder framework~\cite{lu2017knowing,lu2018neural,anderson2018bottom,liu2018context}, \ie, the encoder encodes an image into vector representations and then the decoder decodes them into a language sequence. Since its invention inspired from neural machine translation~\cite{bahdanau2014neural}, this framework has experienced several significant upgrades such as the top-bottom~\cite{xu2015show} and bottom-up~\cite{anderson2018bottom} visual attentions for dynamic encoding, and the reinforced mechanism for sequence decoding~\cite{rennie2017self,gu2017stack,ranzato2015sequence}. However, a ubiquitous problem has never been substantially resolved: when we feed an unseen image scene into the framework, we usually get a simple and trivial caption about the salient objects such as ``there is a dog on the floor'', which is no better than just a list of object detection~\cite{lu2018neural}. This situation is particularly embarrassing in front of the booming ``mid-level'' vision techniques nowadays: we can already detect and segment almost everything in an image~\cite{Hu_2018_CVPR, kirillov2018panoptic, redmon2017yolo9000}. 

We humans are good at telling sentences about a visual scene. Not surprisingly, cognitive evidences~\cite{marr1982vision} show that the visually grounded language generation is not end-to-end and largely attributed to the ``high-level'' symbolic reasoning, that is, once we abstract the scene into symbols, the generation will be almost \emph{disentangled} from the visual perception. For example, as shown in Figure~\ref{fig:1}, from the scene abstraction ``helmet-on-human'' and ``road dirty'', we can say ``a man with a helmet in contryside'' by using the common sense knowledge like ``country road is dirty''. In fact, such collocations and contextual inference in human language can be considered as the \emph{inductive bias} that is apprehended by us from everyday practice, which makes us performing better than machines in high-level reasoning~\cite{lake2017building,battaglia2018relational}. However, the direct exploitation of the inductive bias, \eg, early template/rule-based caption models~\cite{kulkarni2013babytalk, fang2015captions}, is well-known ineffective compared to the encoder-decoder ones, due to the large gap between visual perception and language composition.

In this paper, we propose to incorporate the inductive bias of language generation into the encoder-decoder framework for image captioning, benefiting from the complementary strengths of both symbolic reasoning and end-to-end multi-modal feature mapping. In particular, we use scene graphs~\cite{johnson2018image,N18-1037} to bridge the gap between the two worlds. A scene graph ($\mathcal{G}$) is a unified representation that connects 1) the objects (or entities), 2) their attributes, and 3) their relationships in an image ($\mathcal{I}$) or a sentence ($\mathcal{S}$) by directed edges. Thanks to the recent advances in spatial Graph Convolutional Networks (GCNs)~\cite{marcheggiani2017encoding,li2015gated}, we can embed the graph structure into vector representations, which can be seamlessly integrated into the encoder-decoder. Our key insight is that the vector representations are expected to transfer the inductive bias from the pure language domain to the vision-language domain. 

Specifically, to encode the language prior, we propose the Scene Graph Auto-Encoder (SGAE) that is a sentence self-reconstruction network in the $\mathcal{S}\rightarrow\mathcal{G}\rightarrow\mathcal{D}\rightarrow\mathcal{S}$ pipeline, where $\mathcal{D}$ is a trainable dictionary for the re-encoding purpose of the node features, the $\mathcal{S}\rightarrow\mathcal{G}$ module is a fixed off-the-shelf scene graph language parser~\cite{anderson2016spice}, and the $\mathcal{D}\rightarrow\mathcal{S}$ is a trainable RNN-based language decoder~\cite{anderson2018bottom}. Note that $\mathcal{D}$ is the ``juice'' --- the language inductive bias --- we extract from training SGAE. By sharing $\mathcal{D}$ in the encoder-decoder training pipeline: $\mathcal{I}\rightarrow\mathcal{G}\rightarrow\mathcal{D}\rightarrow\mathcal{S}$, we can incorporate the language prior to guide the end-to-end image captioning. In particular, the $\mathcal{I}\rightarrow \mathcal{G}$ module is a visual scene graph detector~\cite{zellers2018neural} and we introduce a multi-modal GCN for the $\mathcal{G}\rightarrow\mathcal{D}$ module in the captioning pipeline, to complement necessary visual cues that are missing due to the imperfect visual detection.
Interestingly, $\mathcal{D}$ can be considered as a working memory~\cite{vinyals2016matching} that helps to re-key the encoded nodes from $\mathcal{I}$ or $\mathcal{S}$ to a more generic representation with smaller domain gaps. More motivations and the incarnation of $\mathcal{D}$ will be discussed in Section~\ref{subsec:dict}.

We implement the proposed SGAE-based captioning model by using the recently released visual encoder~\cite{ren2015faster} and language decoder~\cite{anderson2018bottom} with RL-based training strategy~\cite{rennie2017self}. Extensive experiments on MS-COCO~\cite{lin2014microsoft} validates the superiority of using SGAE in image captioning. Particularly, in terms of the popular CIDEr-D metric~\cite{vedantam2015cider}, we achieve an absolute 7.2 points improvement over a strong baseline: an upgraded version of Up-Down~\cite{anderson2018bottom}. Then, we advance to a new state-of-the-art \emph{single-model} achieving 127.8 on the Karpathy split and a competitive 125.5 on the official test server even compared to many ensemble models.

In summary, we would like to make the following technical contributions:
\begin{itemize}[leftmargin=.1in]
\item A novel Scene graph Auto-Encoder (SGAE) for learning the feature representation of the language inductive bias.

\item A multi-modal graph convolutional network for modulating scene graphs into visual representations. 

\item A SGAE-based encoder-decoder image captioner with a shared dictionary guiding the language decoding.
\end{itemize}

\section{Related Work}
\noindent\textbf{Image Captioning.}
There is a long history for researchers to develop automatic image captioning methods. Compared with early works which are rules/templates  based~\cite{kuznetsova2012collective,mitchell2012midge,li2011composing}, the modern captioning models have achieved striking advances by three techniques inspired from the NLP field, \ie, encoder-decoder based pipeline~\cite{vinyals2015show}, attention technique~\cite{xu2015show}, and RL-based training objective~\cite{rennie2017self}. Afterwards, researchers tried to discover more semantic information from images and incorporated them into captioning models for better descriptive abilities. For example, some methods exploit object~\cite{lu2018neural}, attribute~\cite{yao2017boosting}, and relationship~\cite{yao2018exploring} knowledge into their captioning models. Compared with these approaches, we use the scene graph as the bridge to integrate object, attribute, and relationship knowledge together to discover more meaningful semantic contexts for better caption generations.

\noindent\textbf{Scene Graphs.}
The scene graph contains the structured semantic information of an image, which includes the knowledge of present objects, their attributes, and pairwise relationships. Thus, the scene graph can provide a beneficial prior for other vision tasks like image retrieval~\cite{johnson2015image}, VQA~\cite{teney2017graph}, and image generation~\cite{johnson2018image}.
By observing the potential of exploiting scene graphs in vision tasks, a variety of approaches are proposed to improve the scene graph generation from images~\cite{zhang2017visual,zellers2018neural,yang2018shuffle,yang2018graph,xu2017scene}. On the another hand, some researchers also tried to extract scene graphs only from textual data~\cite{anderson2016spice,N18-1037}.
In this research, we use~\cite{zellers2018neural} to parse scene graphs from images and~\cite{anderson2016spice} to parse scene graphs from captions.

\noindent\textbf{Memory Networks.}
Recently, many researchers try to augment a working memory into network for preserving a dynamic knowledge base for facilitating subsequent inference~\cite{sukhbaatar2015end,xiong2016dynamic,vinyals2016matching}. Among these methods, differentiable attention mechanisms are usually applied to extract useful knowledge from memory for the tasks on hand. Inspired by these methods, we also implement a memory architecture to preserve humans' inductive bias, guiding our image captioning model to generate more descriptive captions.

\section{Encoder-Decoder Revisited}
\begin{figure}[t]
\centering
\includegraphics[width=1\linewidth,trim = 5mm 3mm 5mm 5mm,clip]{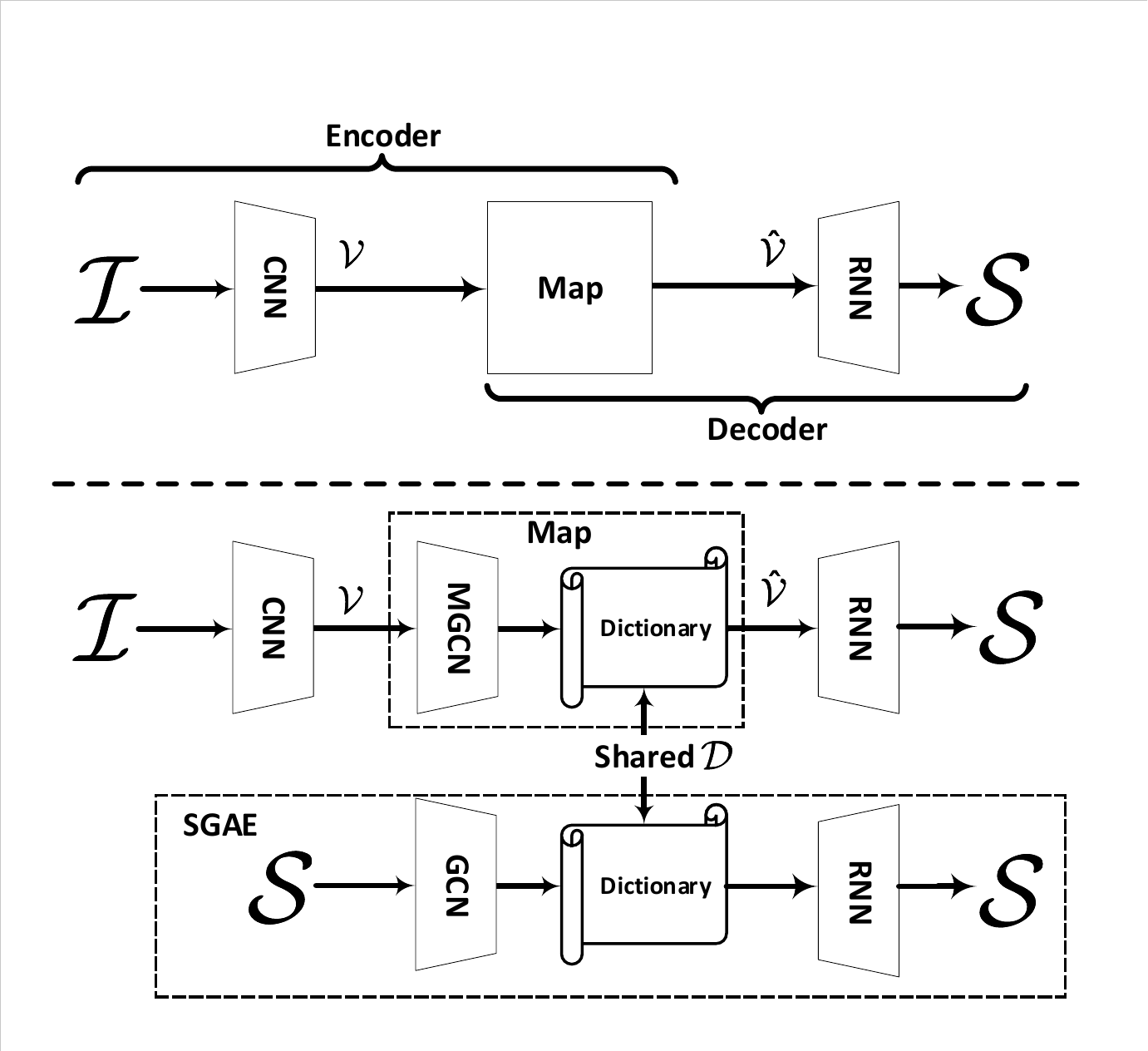}
   \caption{Top: the conventional encoder-decoder. Bottom: our proposed encoder-decoder, where the novel SGAE embeds the language inductive bias in the shared dictionary. }
\label{fig:2}
\end{figure}

As illustrated in Figure~\ref{fig:2}, given an image $\mathcal{I}$, the target of image captioning is to generate a natural language sentence $\mathcal{S}=\{w_1,w_2,...,w_T\}$ describing the image. A state-of-the-art encoder-decoder image captioner can be formulated as:
\begin{equation} \label{equ:equ_ende}
\begin{aligned}
 \textbf{Encoder:} \quad  &\mathcal{V} \leftarrow \mathcal{I}, \\
 \textbf{Map:} \quad  &\hat{\mathcal{V}} \leftarrow \mathcal{V}, \\
 \textbf{Decoder:} \quad & \mathcal{S} \leftarrow \hat{\mathcal{V}}. \\
\end{aligned}
\end{equation}
Usually, an encoder is a Convolutional Neural Network (CNN)~\cite{he2016deep,ren2015faster} that extracts the image feature $\mathcal{V}$; the map is the the widely used attention mechanism~\cite{xu2015show,anderson2018bottom} that re-encodes the visual features into more informative $\hat{\mathcal{V}}$ that is dynamic to language generation; an decoder is an RNN-based language decoder for the sequence prediction of $\mathcal{S}$. Given a ground truth caption $\mathcal{S}^{*}$ for $\mathcal{I}$, we can train this encoder-decoder model by minimizing the cross-entropy loss:
\begin{equation}
    L_{XE} = -\log P(\mathcal{S}^*),
\label{equ:equ_celoss}
\end{equation}
or by maximizing a reinforcement learning (RL) based reward~\cite{rennie2017self} as:
\begin{equation}
    R_{RL} = \mathbb{E}_{\mathcal{S}^s \sim P(\mathcal{S})}[r(\mathcal{S}^s;\mathcal{S}^*)],
    \label{equ:equ_rlloss}
\end{equation}
where $r$ is a sentence-level metric for the sampled sentence $\mathcal{S}^s$ and the ground-truth $\mathcal{S}^*$, \eg, the CIDEr-D~\cite{vedantam2015cider} metric.

This encoder-decoder framework is the core pillar underpinning almost all state-of-the-art image captioners since~\cite{vinyals2015show}. However, it is widely shown brittle to dataset bias~\cite{johnson2017clevr,lu2018neural}. We propose to exploit the language inductive bias, which is beneficial, to confront the dataset bias, which is pernicious, for more human-like image captioning.  As shown in Figure~\ref{fig:2}, the proposed framework is formulated as:
\begin{equation} \label{equ:equ_ende_new}
\begin{aligned}
\textbf{Encoder:} \quad &\mathcal{V} \leftarrow \mathcal{I}, \\
\textbf{Map:} \quad &\hat{\mathcal{V}}\leftarrow R(\mathcal{V}, \mathcal{G};\mathcal{D}),~\mathcal{G} \leftarrow \mathcal{V}, \\
\textbf{Decoder:} \quad & \mathcal{S} \leftarrow \hat{\mathcal{V}}. \\
\end{aligned}
\end{equation}
As can be clearly seen that we focus on modifying the Map module by introducing the scene graph $\mathcal{G}$ into a re-encoder $R$ parameterized by a shared dictionary $\mathcal{D}$. As we will detail in the rest of the paper, we first propose a Scene Graph Auto-Encoder (SGAE) to learn the dictionary $\mathcal{D}$ which embeds the language inductive bias from sentence to sentence self-reconstruction (cf. Section~\ref{sec:ssgae}) with the help of scene graphs. Then, we equip the encoder-decoder with the proposed SGAE to be our overall image captioner (cf. Section~\ref{sec:overall}). Specifically, we use a novel Multi-modal Graph Convolutional Network (MGCN) (cf. Section~\ref{subsec:mmgcn}) to re-encode the image features by using $\mathcal{D}$, narrowing the gap between vision and language.

\section{Auto-Encoding Scene Graphs}
\label{sec:ssgae}
In this section, we will introduce how to learn $\mathcal{D}$ through self-reconstructing sentence $\mathcal{S}$. As shown in Figure~\ref{fig:2}, the process of reconstructing $\mathcal{S}$ is also an encoder-decoder pipeline. Thus, by slightly abusing the notations, we can formulate SGAE as:
\begin{equation} \label{equ:equ_ende_sen}
\begin{aligned}
\textbf{Encoder:} \quad &\mathcal{X}\leftarrow \mathcal{G} \leftarrow \mathcal{S}, \\
\textbf{Map:} \quad  &\hat{\mathcal{X}} \leftarrow R(\mathcal{X};\mathcal{D}), \\
\textbf{Decoder:} \quad & \mathcal{S} \leftarrow \hat{\mathcal{X}}. \\
\end{aligned}
\end{equation}
Next, we will detail every component mentioned in Eq.~\eqref{equ:equ_ende_sen}. 

\subsection{Scene Graphs}
\label{subsec:sgf}
We introduce how to implement the step $\mathcal{G}\leftarrow \mathcal{S}$, \ie, from sentence to scene graph. Formally, a scene graph is a tuple $\mathcal{G}=(\mathcal{N},\mathcal{E})$, where $\mathcal{N}$ and $\mathcal{E}$ are the sets of nodes and edges, respectively. There are three kinds of nodes in $\mathcal{N}$: object node $o$, attribute node $a$, and relationship node $r$. We denote $o_i$ as the $i$-th object, $r_{ij}$ as the relationship between object $o_i$ and $o_j$, and $a_{i,l}$ as the $l$-th attribute of object $o_i$.
For each node in $\mathcal{N}$, it is represented by a $d$-dimensional vector, \ie, $\bm{e}_o$, $\bm{e}_a$, and $\bm{e}_r$. In our implementation, $d$ is set to $1,000$. In particular, the node features are trainable label embeddings. The edges in $\mathcal{E}$ are formulated as follows:
\begin{itemize}[nolistsep]
\item if an object $o_i$ owns an attribute $a_{i,l}$, assigning a directed edge from $o_i$ to $a_{i,l}$;
\item if there is one relationship triplet $<o_i-r_{ij}-o_j>$ appears, assigning two directed edges from $o_i$ to $r_{ij}$ and from $r_{ij}$ to $o_j$, respectively.
\end{itemize}
Figure~\ref{fig:fig_gnn} shows one example of $\mathcal{G}$, which contains $7$ nodes in $\mathcal{N}$ and $6$ directed edges in $\mathcal{E}$.

We use the scene graph parser provided by~\cite{anderson2016spice} for scene graphs $\mathcal{G}$ from sentences, where a syntactic dependency tree is built by~\cite{klein2003accurate} and then a rule-based method~\cite{schuster2015generating} is applied for transforming the tree to a scene graph.

\subsection{Graph Convolution Network}
\label{subsec:gnn}
\begin{figure}[t]
\centering
\includegraphics[width=1\linewidth,trim = 5mm 3mm 5mm 5mm,clip]{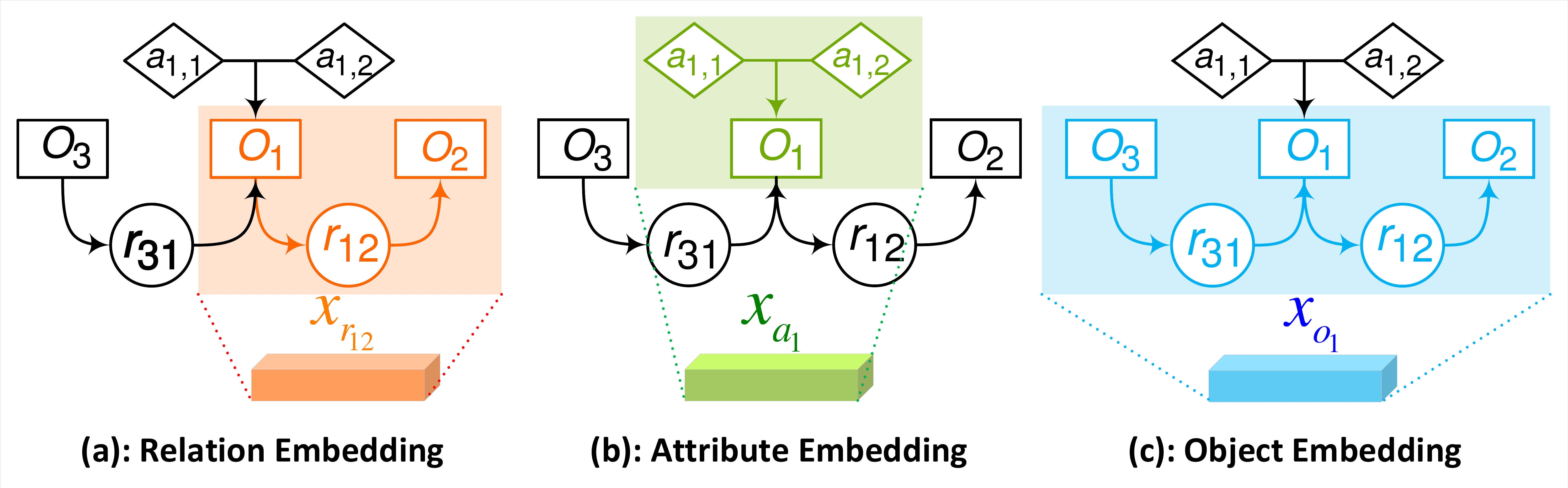}
   \caption{Graph Convolutional Network. In particular, it is spatial convolution, where the colored neighborhood is ``convolved'' for the resultant embedding.}
\label{fig:fig_gnn}
\end{figure}
We present the implementation for the step $\mathcal{X}\leftarrow\mathcal{G}$ in Eq.~\eqref{equ:equ_ende_sen}, \ie, how to transform the original node embeddings $\bm{e}_o$, $\bm{e}_a$, and $\bm{e}_r$ into a new set of context-aware embeddings $\mathcal{X}$. Formally, $\mathcal{X}$ contains three kinds of $d$-dimensional embeddings: relationship embedding $\bm{x}_{r_{ij}}$ for relationship node $r_{ij}$, object embedding $\bm{x}_{o_i}$ for object node $o_i$, and attribute embedding $\bm{x}_{a_i}$ for object node $o_i$. In our implementation, $d$ is set to $1,000$. We use four \emph{spatial graph convolutions}: $g_r$, $g_a$, $g_s$, and $g_o$ for generating the above mentioned three kinds of embeddings. In our implementation, all these four functions have the same structure with independent parameters: a vector concatenation input to a fully-connected layer, followed by an ReLU.

\noindent\textbf{Relationship Embedding $\mathbf{x}_{r_{ij}}$:}
Given one relationship triplet $<o_i-r_{ij}-o_j>$ in $\mathcal{G}$, we have:
\begin{equation}
\label{equ:equ_gnnr}
    \bm{x}_{r_{ij}}=g_r(\bm{e}_{o_i}, \bm{e}_{r_{ij}}, \bm{e}_{o_j}),
\end{equation}
where the context of a relationship triplet is incorporated together. Figure~\ref{fig:fig_gnn} (a) shows such an example.

\noindent\textbf{Attribute Embedding $\bm{x}_{a_i}$:}
Given one object node $o_i$ with all its attributes $a_{i,1:Na_i}$ in $\mathcal{G}$, where $Na_i$ is the number of attributes that the object $o_i$ has, then $\bm{x}_{{a}_i}$ for $o_i$ is:
\begin{equation}
   \bm{x}_{a_i} =\frac{1}{Na_i}\sum_{l=1}^{Na_i}g_a(\bm{e}_{o_i},\bm{e}_{a_{i,l}}),
\label{equ:equ_gnna}
\end{equation}
where the context of this object and all its attributes are incorporated. Figure~\ref{fig:fig_gnn} (b) shows such an example.

\noindent\textbf{Object Embedding $\bm{x}_{o_i}$:}
In $\mathcal{G}$, $o_i$ can act as ``subject'' or ``object'' in relationships, which means $o_i$ will play different roles due to different edge directions. Then, different functions should be used to incorporate such knowledge. For avoiding ambiguous meaning of the same ``predicate'' in different context, knowledge of the whole relationship triplets where $o_i$ appears should be incorporated into $\bm{x}_{o_i}$. One simple example for ambiguity is that, in $<$hand-with-cup$>$, the predicate ``with'' may mean ``hold'', while in $<$head-with-hat$>$, ``with'' may mean ``wear''. Therefore, $\bm{x}_{o_i}$ can be calculated as:
\begin{equation}
\label{equ:equ_gnno}
\begin{split}
    &\bm{x}_{o_i} = \frac{1}{Nr_i}[\sum_{o_j\in sbj(o_i)}g_s(\bm{e}_{o_i},\bm{e}_{o_j},\bm{e}_{r_{ij}}) \\
    &+\sum_{o_k\in obj(o_i)}g_o(\bm{e}_{o_k},\bm{e}_{o_i},\bm{e}_{r_{ki}})].
\end{split}
\end{equation}
For each node $o_j \in sbj(o_i)$, it acts as ``object'' while $o_i$ acts as ``subject'', \eg, $sbj(o_1)=\{o_2\}$ in Figure~\ref{fig:fig_gnn} (c). $Nr_i=|sbj(i)|+|obj(i)|$ is the number of relationship triplets where $o_i$ is present. Figure~\ref{fig:fig_gnn} (c) shows this example.

\subsection{Dictionary}
\label{subsec:dict}
\begin{figure}[t]
\centering
\includegraphics[width=1\linewidth,trim = 5mm 3mm 5mm 5mm,clip]{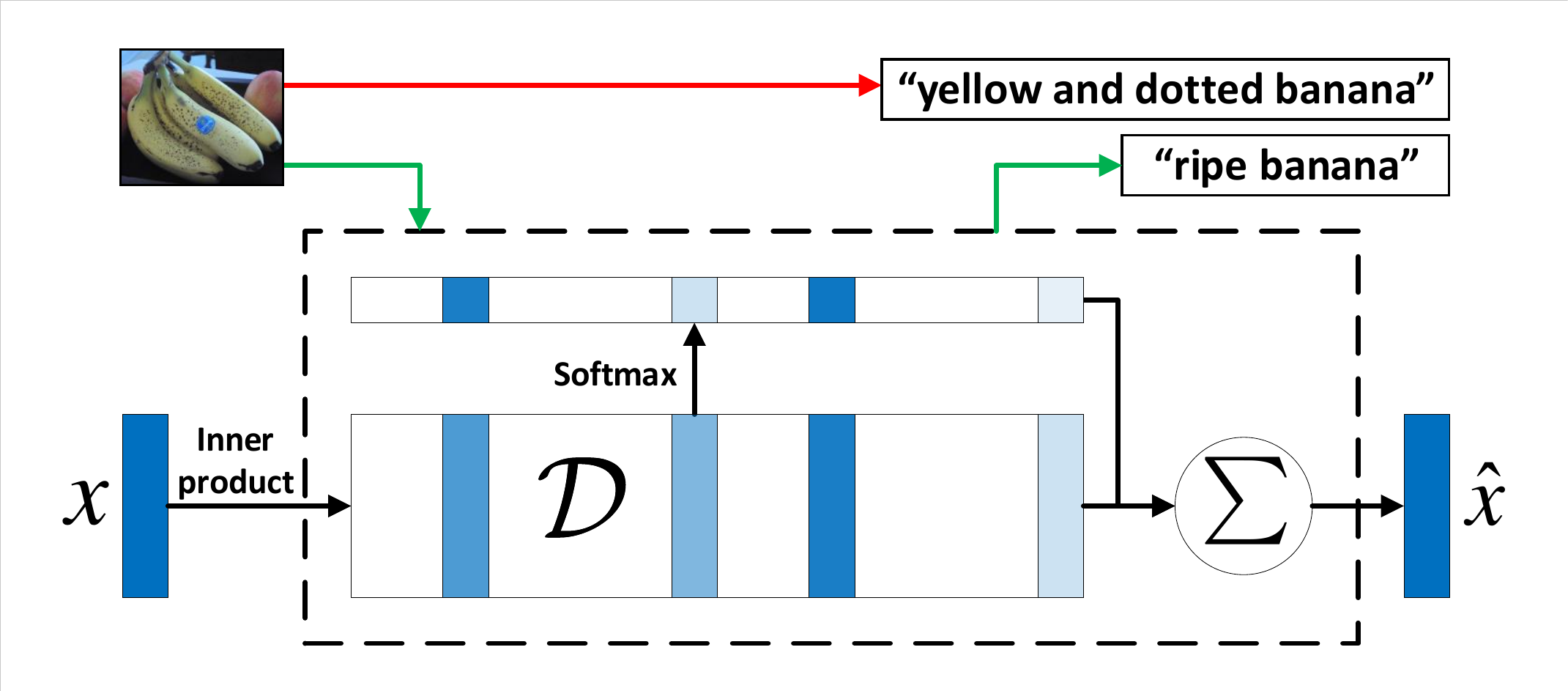}
   \caption{The visualization of the re-encoder function $R$. The black dashed block shows the operation of re-encoding. The top part demonstrates how ``imagination'' is achieved by re-encoding: green line shows the generated phrase by re-encoding, while the red line shows the one without re-encoding.}
\label{fig:fig_dict}
\end{figure}
Now we introduce how to learn the dictionary $\mathcal{D}$ and then use it to re-encode $\hat{\mathcal{X}} \leftarrow R(\mathcal{X};\mathcal{D})$ in Eq.~\eqref{equ:equ_ende_sen}. Our key idea is inspired by using the working memory to preserve a dynamic knowledge base for run-time inference, which is widely used in textual QA~\cite{sukhbaatar2015end}, VQA~\cite{xiong2016dynamic}, and one-shot classification~\cite{vinyals2016matching}. Our $\mathcal{D}$ aims to embed language inductive bias in language composition. Therefore, we propose to place the dictionary learning into the sentence self-reconstruction framework. Formally, we denote $\mathcal{D}$ as a $d \times K$ matrix $\bm{D} = \{\bm{d}_1,\bm{d}_2,...,\bm{d}_K\}$. The $K$ is set as $10,000$ in implementation. Given an embedding vector $\bm{x} \in \mathcal{X}$, the re-encoder function $R_\mathcal{D}$ can be formulated as:
\begin{equation} \label{equ:equ_recre}
\hat{\bm{x}}=R(\bm{x};\mathcal{D})= \bm{D} \bm{\alpha}=\sum_{k=1}^K\alpha_k\bm{d}_k,
\end{equation}
where $\bm{\alpha} = \text{softmax}(\bm{D}^T\bm{x})$ can be viewed as the ``key'' operation in memory network~\cite{sukhbaatar2015end}.
As shown in Figure~\ref{fig:fig_dict}, this re-encoding offers some interesting ``imagination'' in human common sense reasoning. For example, from ``yellow and dotted banana'', after re-encoding, the feature will be more likely to generate ``ripe banana''. 

We deploy the attention structure in~\cite{anderson2018bottom} for reconstructing $\mathcal{S}$. Given a reconstructed $\mathcal{S}$, we can use the training objective in Eq.~\eqref{equ:equ_celoss} or~\eqref{equ:equ_rlloss} to train SGAE parameterized by $\mathcal{D}$ in an end-to-end fashion. Note that training SGAE is unsupervised, that is, SGAE offers a potential never-ending learning from large-scale unsupervised inductive bias learning for $\mathcal{D}$. Some preliminary studies are reported in Section~\ref{subsec:exp_web}. 

\section{Overall Model: SGAE-based Encoder-Decoder}
\label{sec:overall}
In this section, we will introduce the overall model: SGAE-based Encoder-Decoder as sketched in Figure~\ref{fig:2} and Eq.~\eqref{equ:equ_ende_new}.
\subsection{Multi-modal Graph Convolution Network}
\label{subsec:mmgcn}
The original image features extracted by CNN are not ready for use for the dictionary re-encoding as in Eq.~\eqref{equ:equ_recre}, due to the large gap between vision and language. To this end, we propose a Multi-modal Graph Convolution Network (MGCN) to first map the visual features $\mathcal{V}$ into a set of scene graph-modulated features ${\mathcal{V}}'$.

Here, the scene graph $\mathcal{G}$ is extracted by an image scene graph parser that contains an object proposal detector, an attribute classifier, and a relationship classifier. In our implementation, we use Faster-RCNN as the object detector~\cite{ren2015faster}, MOTIFS relationship detector~\cite{zellers2018neural} as the relationship classifier, and we use our own attribute classifier: an small fc-ReLU-fc-Softmax network head. The key representation difference between the sentence-parsed $\mathcal{G}$ and the image-parsed $\mathcal{G}$ is that the node $o_i$ is not only the label embedding. In particular, we use the RoI features pre-trained from Faster RCNN and then fuse the detected label embedding $\bm{e}_{o_i}$ with the visual feature $\bm{v}_{o_i}$, into a new node feature $\bm{u}_{o_i}$:
\begin{equation}\label{eq:fuse}
    \bm{u}_{o_i} = \text{ReLU}(\bm{W}_1 \bm{e}_{o_i} + \bm{W}_2 \bm{v}_{o_i}) - (\bm{W}_1 \bm{e}_{o_i} - \bm{W}_2 \bm{v}_{o_i})^2.
\end{equation}
where $\bm{W}_1$ and $\bm{W}_2$ are the fusion parameters following~\cite{zhang2018learning}. Compared to the popular bi-linear fusion~\cite{zhang2018learning}, Eq~\eqref{eq:fuse} is empirically shown a faster convergence of training the label embeddings in our experiments. The rest node embeddings: $\bm{u}_{r_{ij}}$ and $\bm{u}_{a_i}$ are obtained in a similar way. The differences between two scene graphs generated from $\mathcal{I}$ and $\mathcal{S}$ are visualized in Figure~\ref{fig:1}, where the image $\mathcal{G}$ is usually more simpler and nosier than the sentence $\mathcal{G}$.

Similar to the GCN used in Section~\ref{subsec:gnn}, MGCN also has an ensemble of four functions $f_r$, $f_a$, $f_s$ and $f_o$, each of which is a two-layer structure: fc-ReLU with independent parameters. And the computation of relationship, attribute and object embeddings are similar to Eq.~\eqref{equ:equ_gnnr}, Eq.~\eqref{equ:equ_gnna}, and Eq.~\eqref{equ:equ_gnno}, respectively. After computing $\mathcal{V}'$ by using MGCN, we can adopt Eq.~\eqref{equ:equ_recre} to re-encode $\mathcal{V}'$ as $\hat{\mathcal{V}}$ and feed $\hat{\mathcal{V}}$ to the decoder for generating language $\mathcal{S}$. In particular, we deploy the attention structure in~\cite{anderson2018bottom} for the generation.

\subsection{Training and Inference}
\label{subsec:tai}
Following the common practice in deep-learning feature transfer~\cite{devlin2018bert,yosinski2014transferable}, we use the SGAE pre-trained $\mathcal{D}$ as the initialization for the $\mathcal{D}$ in our overall encoder-decoder for image captioning. In particular, we intentionally use a very small learning rate (\eg, $10^{-5}$) for fine-tuning $\mathcal{D}$ to impose the sharing purpose. The overall training loss is hybrid: we use the cross-entropy loss in Eq.~\eqref{equ:equ_celoss} for $20$ epochs and then use the RL-based reward in Eq.~\eqref{equ:equ_rlloss} for another $40$ epochs. 

For inference in language generation, we adopt the beam search strategy~\cite{rennie2017self} with a beam size of 5.

\section{Experiments}
\subsection{Datasets, Settings, and Metrics}
\noindent\textbf{MS-COCO~\cite{lin2014microsoft}.} There are two standard splits of MS-COCO: the official online test split and the 3rd-party Karpathy split~\cite{karpathy2015deep} for offline test. The first split has  $82,783/40,504/40,775$ train/val/test images, each of which has 5 human labeled captions. The second split has $113,287/5,000/5,000$ train/val/test images, each of which has 5 captions.

\noindent\textbf{Visual Genome~\cite{krishna2017visual} (VG).} This dataset has abundant scene graph annotations, \eg, objects' categories, objects' attributes, and pairwise relationships, which can be exploited to train the object proposal detector, attribute classifier, and relationship classifier~\cite{zellers2018neural} as our image scene graph parser.

\noindent\textbf{Settings.}
For captions, we used the following steps to pre-process the captions: we first tokenized the texts on white space; then we changed all the words to lowercase; we also deleted the words which appear less than $5$ times; at last, we trimmed each caption to a maximum of $16$ words. This results in a vocabulary of $10,369$ words. This pre-processing was also applied in VG. It is noteworthy that except for ablative studies, these additional text descriptions from VG were not used for training the captioner. Since the object, attribute, and relationship annotations are very noisy in VG dataset, we filter them by keeping the objects, attributes, and relationships which appear more than $2,000$ times in the training set. After filtering, the remained $305$ objects, $103$ attributes, and $64$ relationships are used to train our object detector, attribute classifier and relationship classifier.

We chose the language decoder proposed in~\cite{anderson2018bottom}. The number of hidden units of both LSTMs used in this decoder is set to $1000$. For training SGAE in Eq.~\eqref{equ:equ_ende_sen}, the decoder is firstly set as $\mathcal{S} \leftarrow \mathcal{X}$ and $\mathcal{D}$ is not trained to learn a rudiment encoder and decoder. We used the corss-entropy loss in Eq.~\eqref{equ:equ_celoss} to train them for $20$ epochs. Then the decoder was set as $\mathcal{S} \leftarrow \hat{\mathcal{X}}$ to train $\mathcal{D}$ by cross-entropy loss for another $20$ epochs. The learning rate was initialized to $5e^{-4}$ for all parameters and we decayed them by $0.8$ for every $5$ epochs. For training our SGAE-based encoder-decoder, we followed Eq.~\eqref{equ:equ_ende_new} to generate $\mathcal{S}$ with shared $\mathcal{D}$ pre-trained from SGAE. The decoder was set as $\mathcal{S} \leftarrow \{\hat{\mathcal{V}}, \mathcal{V}'\}$, where $\mathcal{V}'$ and $\hat{\mathcal{V}}$ can provide visual clues and high-level semantic contexts respectively. In this process, cross-entropy loss was first used to train the network for $20$ epochs and then the RL-based reward was used to train for another $40$ epochs. The learning rate for $\mathcal{D}$ was initialized to $5e^{-5}$ and for other parameters it was $5e^{-4}$, and all these learning rates were decayed by $0.8$ for every $5$ epochs. Adam optimizer~\cite{kingma2014adam} was used for batch size $100$.

\noindent\textbf{Metrics.}
We used four standard automatic evaluations metrics: CIDEr-D~\cite{vedantam2015cider}, BLEU~\cite{papineni2002bleu}, METEOR\cite{banerjee2005meteor} and ROUGE~\cite{lin2004rouge}.

\subsection{Ablative Studies}
We conducted extensive ablations for architecture (Section~\ref{subsec:arch}), language corpus (Section~\ref{subsec:exp_web}), and sentence reconstruction quality (Section~\ref{subsec:sg_qual}). For simplicity, we use \textbf{SGAE} to denote our SGAE-based encoder-decoder captioning model.

\begin{table}[t]
\begin{center}
\caption{The performances of various methods on MS-COCO Karpathy split. The metrics: B@N, M, R, C and S denote BLEU@N, METEOR, ROUGE-L, CIDEr-D and SPICE. Note that the ${fuse}$ subscript indicates fused models while the rest methods are all single models. The best results for each metric on fused models and single models are marked in boldface separately.}
\label{table:tab_kap}
\scalebox{0.78}{
\begin{tabular}{l c c c c c c}
		\hline
		   Models   & B@1 & B@4 & M & R &  C & S\\ \hline
           SCST~\cite{rennie2017self}       & $-$  & $34.2$ & $26.7$ & $55.7$ & $114.0$ & $-$   \\ 
           LSTM-A~\cite{yao2017boosting}   & $78.6$  & $35.5$ & $27.3$ & $56.8$ & $118.3$ & $20.8$   \\ 
           StackCap~\cite{gu2017stack}    & $78.6$  & $36.1$ & $27.4$ & $-$ & $120.4$ & $-$   \\ 
           Up-Down~\cite{anderson2018bottom}  & $79.8$ & $36.3$ & $27.7$ & $56.9$ & $120.1$ & $21.4$ \\ 
           CAVP~\cite{liu2018context}  & $-$& $\mathbf{38.6}$ & $28.3$ & $58.5$ & $126.3$ & $21.6$ \\ 
           GCN-LSTM${^\dagger}$~\cite{yao2018exploring}  & $80.0$  & $37.1$ & $28.0$ & $57.3$ & $122.8$ & $21.1$ \\ 
           GCN-LSTM~\cite{yao2018exploring}  & $80.5$  & $38.2$ & $\mathbf{28.5}$ & $58.3$ & $127.6$ & $22.0$ \\ \hline
           Base       & $79.9$ & $36.8$& $27.7$ & $57.0$ & $120.6$ & $20.9$\\ 
           Base+MGCN       & $80.2$ & $37.2$& $27.9$ & $57.5$ & $123.4$ & $21.2$\\ 
           Base+$\bm{D}$ w/o GCN       & $80.2$ & $37.3$& $27.8$ & $58.0$ & $124.2$ & $21.4$\\ 
           Base+$\bm{D}$       & $80.4$ & $37.7$& $28.1$ & $58.2$ & $125.7$ & $21.4$\\ 
           SGAE    & $\mathbf{80.8}$ & $38.4$ & $28.4$ & $\mathbf{58.6}$ & $\mathbf{127.8}$ & $\mathbf{22.1}$\\ \hline 
           SGAE$_{fuse}$    & $\mathbf{81.0}$ & $\mathbf{39.0}$ & $28.4$ & $\mathbf{58.9}$ & $\mathbf{129.1}$ & $\mathbf{22.2}$\\ 
           GCN-LSTM$_{fuse}$~\cite{yao2018exploring} & $80.9$  & $38.3$ & $\mathbf{28.6}$ & $58.5$ & $128.7$ & $22.1$ \\ \hline
\end{tabular}
}
\end{center}
\end{table}
\subsubsection{Architecture}
\label{subsec:arch}

\noindent\textbf{Comparing Methods}.
For quantifying the importance of the proposed GCN, MGCN, and dictionary $\mathcal{D}$, we ablated our SGAE with the following baselines: \textbf{Base:} We followed the pipeline given in Eq~\eqref{equ:equ_ende} without using GCN, MGCN, and $\mathcal{D}$. This baseline is the benchmark for other ablative baselines.
\textbf{Base+MGCN:} We added MGCN to compute the multi-modal embedding set $\hat{\mathcal{V}}$. This baseline is designed for validating the importance of MGCN.
\textbf{Base+}$\bm{D} \textbf{ w/o GCN}$: We learned $\mathcal{D}$ by using Eq.~\eqref{equ:equ_ende_sen}, while GCN is not used and only word embeddings of $\mathcal{S}$ were input to the decoder. Also, MGCN in Eq.~\eqref{equ:equ_ende_new} is not used. This baseline is designed for validating the importance of GCN.
\textbf{Base+}$\bm{D}$\textbf{:} Compared to Base, we learned $\mathcal{D}$ by using GCN. And MGCN in Eq.~\eqref{equ:equ_ende_new} was not used. This baseline is designed for validating the importance of the shared $\mathcal{D}$.

\noindent\textbf{Results.} The middle section of Table~\ref{table:tab_kap} shows the performances of the ablative baselines on MS-COCO Karpathy split. Compared with Base, our SGAE can boost the CIDEr-D by absolute $7.2$. By comparing Base+MGCN, Base+$\bm{D}$ w/o GCN, and Base+$\bm{D}$ with Base, we can find that all the performances are improved, which demonstrate that the proposed MGCN, GCN, and $\mathcal{D}$ are all indispensable for advancing the performances. We can also observe that the performances of Base+$\bm{D}$ or Base+$\bm{D}$ w/o GCN are better than Base+MGCN, which suggests that the language inductive bias plays an important role in generating better captions.

\begin{figure*}[t]
\centering
\includegraphics[width=1\linewidth,trim = 5mm 3mm 5mm 5mm,clip]{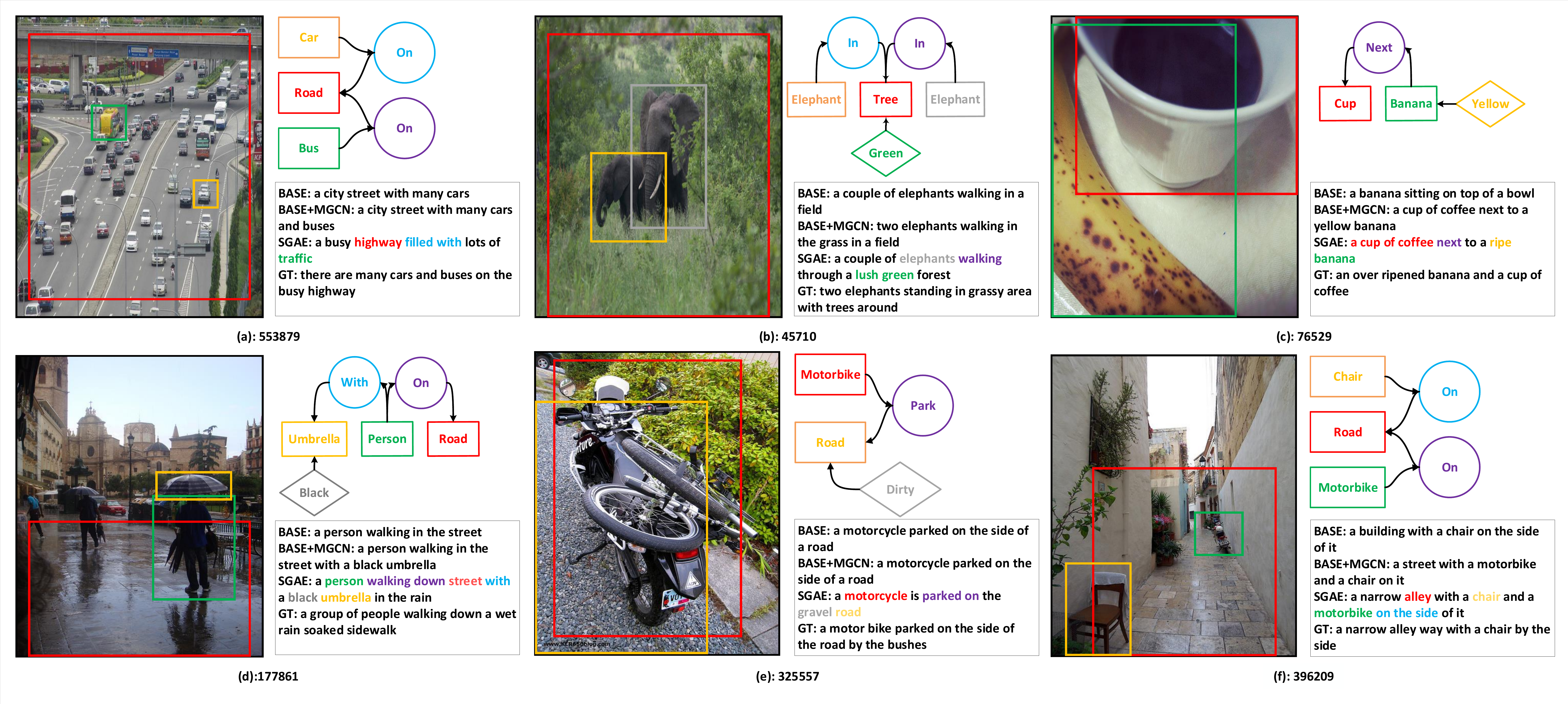}
   \caption{Qualitative examples of different baselines. For each figure, the image scene graph is pruned to avoid clutter.  The id refers to the image id in MS-COCO. Word colors correspond to nodes in the detected scene graphs. }
   \vspace{-0.1in}
\label{fig:fig_exp}
\end{figure*}
\begin{figure*}[t]
\centering
\includegraphics[width=1\linewidth,trim = 5mm 3mm 5mm 5mm,clip]{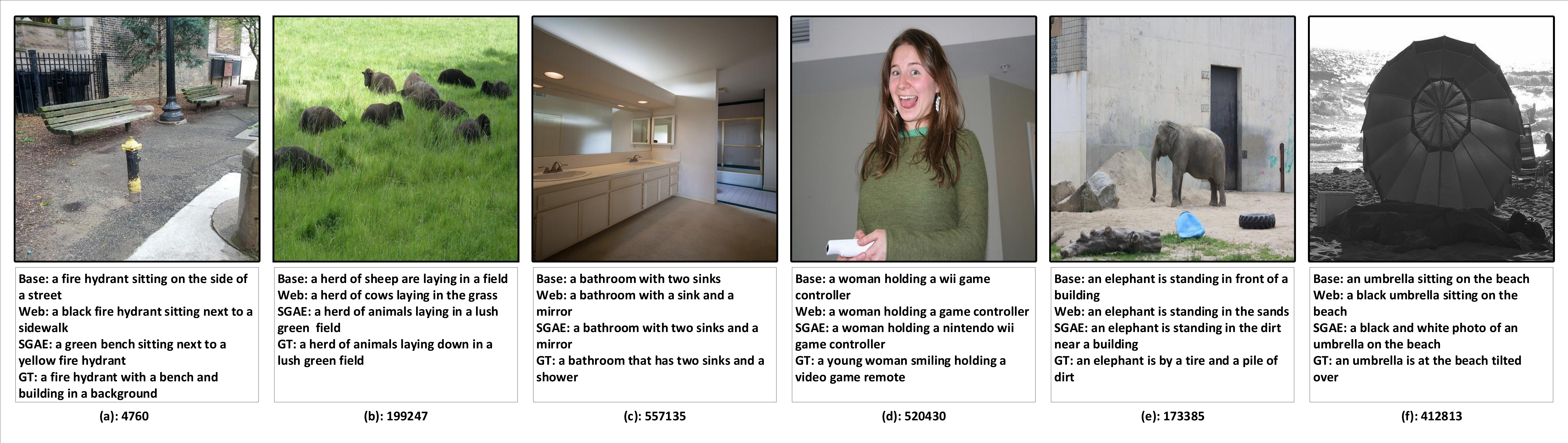}
   \caption{Captions generated by using different language corpora. }
   \vspace{-0.1in}
\label{fig:fig_exp2}
\end{figure*}
\noindent\textbf{Qualitative Examples.}
Figure~\ref{fig:fig_exp} shows $6$ examples of the generated captions using different baselines. We can see that compared with captions generated by Base, Base+MGCN's descriptions usually contain more descriptions about objects' attributes and pairwise relationships. For captions generated by SGAE, they are more complex and descriptive. For example, in Figure~\ref{fig:fig_exp} (a), the word ``busy'' will be used to describe the heavy traffic; in (b) the scene ``forest'' can be deduced from ``trees''; and in (d), the weather ``rain'' will be inferred from ``umbrella'.

\begin{table}[t]
\begin{center}
\caption{The performances of using different language corpora}
\label{table:tab_web}
\scalebox{0.95}{
\begin{tabular}{l c c c c c c }
		\hline
		   Models   & B@1 & B@4 & M & R &  C & S\\ \hline
		   Base       & $79.9$ & $36.8$& $27.7$ & $57.0$ & $120.6$ & $20.9$\\ 
		   Web      & $80.2$ & $37.8$& $28.0$ & $58.2$ & $123.2$ & $21.3$\\ 
		   SGAE     & $\mathbf{80.8}$ & $\mathbf{38.4}$ & $\mathbf{28.4}$ & $\mathbf{58.6}$ & $\mathbf{127.8}$ & $\mathbf{22.1}$\\ \hline
\end{tabular}
}
\end{center}
\end{table}

\begin{table}[t]
\begin{center}
\caption{The performances of using different scene graphs}
\label{table:tab_orac}
\scalebox{0.95}{
\begin{tabular}{l c c c c c c }
		\hline
		   Models   & B@1 & B@4 & M & R &  C & S\\ \hline
            
           $\widehat{\mathcal{X}}$  & $90.3$ & $53.8$& $34.3$ & $66.5$ & $153.2$ & $30.6$\\ 
           $\mathcal{X}$  & $\mathbf{93.9}$ & $\mathbf{65.2}$& $\mathbf{38.5}$ & $\mathbf{71.8}$ & $\mathbf{177.0}$ & $\mathbf{34.3}$\\ \hline
           SGAE    & $80.8$ & $38.4$ & $28.4$ & $58.6$ & $127.8$ & $22.1$\\ \hline
\end{tabular}
}
\end{center}
\end{table}

\begin{table*}[t]
\begin{center}
\caption{The performances of various methods on the online MS-COCO test server. The metrics: B@N, M, R, and C denote BLEU@N, METEOR, ROUGE-L, and CIDEr-D.}
\label{table:tab_online}
\scalebox{0.86}{
\begin{tabular}{l c c c c c c c c c c c c c c}
		\hline
		 Model & \multicolumn{2}{c}{B@1}  & \multicolumn{2}{c}{B@2}  & \multicolumn{2}{c}{B@3} & \multicolumn{2}{c}{B@4} &\multicolumn{2}{c}{M} &\multicolumn{2}{c}{R-L} & \multicolumn{2}{c}{C-D}\\ \hline 
		 Metric  &  c5 & c40 & c5 & c40 &  c5 & c40 & c5 & c40 & c5 & c40 & c5 & c40 & c5 & c40 \\ \hline
           SCST~\cite{rennie2017self}      & $78.1$ & $93.7$ & $61.9$ & $86.0$ & $47.0$ & $75.9$ & $35.2$ & $64.5$ & $27.0$ & $35.5$ & $56.3$ & $70.7$ & $114.7$ & $116.0$  \\
           LSTM-A~\cite{yao2017boosting}   & $78.7$ & $93.7$ & $62.7$ & $86.7$ & $47.6$ & $76.5$ & $35.6$ & $65.2$ & $27.0$ & $35.4$ & $56.4$ & $70.5$ & $116.0$ & $118.0$  \\ 
           StackCap~\cite{gu2017stack}     & $77.8$ & $93.2$ & $61.6$ & $86.1$ & $46.8$ & $76.0$ & $34.9$ & $64.6$ & $27.0$ & $35.6$ & $56.2$ & $70.6$ & $114.8$ & $118.3$   \\ 
           Up-Down~\cite{anderson2018bottom}    & $80.2$ & $95.2$ & $64.1$ & $88.8$ & $49.1$ & $79.4$ & $36.9$ & $68.5$ & $27.6$ & $36.7$ & $57.1$ & $72.4$ & $117.9$& $120.5$  \\ 
           CAVP~\cite{liu2018context}           & $80.1$ & $94.9$ & $64.7$ & $88.8$ & $50.0$ & $79.7$ & $37.9$ & $69.0$ & $28.1$ & $37.0$ & $58.2$ & $73.1$ & $121.6$& $123.8$    \\
           SGAE$_{single}$    & $80.6$ & $95.0$  & $65.0$ & $88.9$ & $50.1$  & $79.6$ & $37.8$ & $68.7$& $28.1$ & $37.0$ & $58.2$ & $73.1$ & $122.7$ & $125.5$\\ 
           SGAE$_{fuse}$    & $\mathbf{81.0}$ & $\mathbf{95.3}$  & $\mathbf{65.6}$ & $\mathbf{89.5}$ & $\mathbf{50.7}$  & $\mathbf{80.4}$ & $\mathbf{38.5}$ & $\mathbf{69.7}$& $\mathbf{28.2}$ & $\mathbf{37.2}$ & $\mathbf{58.6}$ & $\mathbf{73.6}$ & $\mathbf{123.8}$ & $\mathbf{126.5}$\\ 
           \hline
\end{tabular}
}
\end{center}
\vspace{-0.1in}
\end{table*}

\begin{figure}[t]
\centering
\includegraphics[width=1\linewidth]{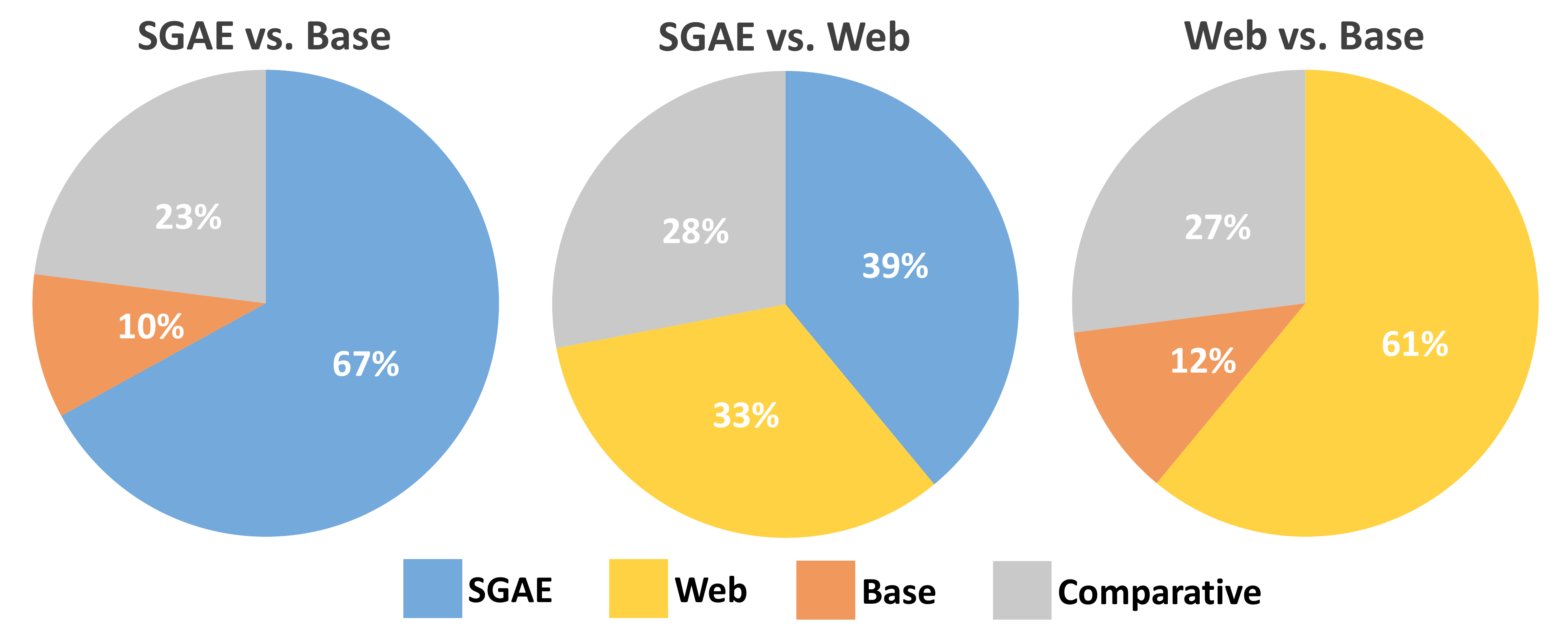}
   \caption{The pie charts each comparing the two methods in human evaluation. Each color indicates the percentage of users who consider that the corresponding method generates more descriptive captions. In particular, the gray color indicates that the two methods are comparative.}
   \vspace{-0.1in}
\label{fig:fig_pie}
\end{figure}
\subsubsection{Language Corpus}
\label{subsec:exp_web}
\noindent\textbf{Comparing Methods}.
To test the potential of using large-scale corpus for learning a better $\mathcal{D}$, we used the texts provided by VG instead of MS-COCO to learn $\mathcal{D}$, and then share the learned $\mathcal{D}$ in the encoder-decoder pipeline. The results are demonstrated in Table~\ref{table:tab_web}, where \textbf{Web} means results obtained by using sentences from VG.

\noindent\textbf{Results.} We can observe that by using the web description texts, the performances of generated captions are boosted compared with Base, which validates the potential of our proposed model in exploiting additional Web texts. We can also see that by using texts provided by MS-COCO itself (SGAE), the generated captions have better scores than using Web texts. This is intuitively reasonable since $\mathcal{D}$ can preserve more useful clues when a matched language corpus is given. Both of these two comparisons validate the effectiveness of $\mathcal{D}$ in two aspects: $\mathcal{D}$ can memorize common inductive bias from the additional unmatched Web texts or specific inductive bias from a matched language corpus.

\noindent\textbf{Qualitative Examples.}
Figure~\ref{fig:fig_exp2} shows $6$ examples of generated captions by using different language corpora. Generally, compared with captions generated by Base, the captions of Web and SGAE are more descriptive. Specifically, the captions generated by using the matched language corpus can usually describe a scene by some specific expressions in the dataset, while more general expressions will appear in captions generated by using Web texts. For example, in Figure~\ref{fig:fig_exp2} (b), SGAE uses ``lush green field'' as GT captions while Web uses ``grass'' ; or in (e), SGAE prefers ``dirt'' while Web prefers ``sand''.

\noindent\textbf{Human Evaluation.}
For better evaluating the qualities of the generated captions by using different language corpora, we conducted human evaluation with $30$ workers. We showed them two captions generated by different methods and asked them which one is more descriptive. For each pairwise comparison, $100$ images are randomly extracted from the Karpathy split for them to compare. The results of the comparisons are shown in Figure~\ref{fig:fig_pie}. From these pie charts, we can observe that when a $\mathcal{D}$ is used, the generated captions are evaluated to be more descriptive.

\subsubsection{Sentence Reconstruction}
\label{subsec:sg_qual}
\noindent\textbf{Comparing Methods}.
We investigated how well the sentences are reconstructed in training SGAE in Eq.~\eqref{equ:equ_ende_sen}, with or without using the re-encoding by $\mathcal{D}$, that is, we denote $\widehat{\mathcal{X}}$ as the pipeline using $\mathcal{D}$ and $\mathcal{X}$ as the pipeline directly reconstructing sentences from their scene graph node features. Such results are given in Table~\ref{table:tab_orac}.

\noindent\textbf{Analysis.}
As we can see, the performances of using direct scene graph features $\widehat{\mathcal{X}}$ are much better than those ($\mathcal{X}$) imposed with $\mathcal{D}$ for re-encoding. This is reasonable since $\mathcal{D}$ will regularize the reconstruction and thus encourages the learning of language inductive bias. Interestingly, the gap between $\hat{\mathcal{X}}$ and SGAE suggest that we should develop a more powerful image scene graph parser for improving the quality of $\mathcal{G}$ in Eq.~\eqref{equ:equ_ende_new}, and a stronger re-encoder should be designed for extracting more preserved inductive bias when only low-quality visual scene graphs are available.

\subsection{Comparisons with State-of-The-Arts}

\noindent\textbf{Comparing Methods.} Though there are various captioning models developed in recent years, for fair comparison, we only compared SGAE with some encoder-decoder methods trained by the RL-based reward (Eq.~\eqref{equ:equ_rlloss}), due to their superior performances. Specifically, we compared our methods with \textbf{SCST}~\cite{rennie2017self}, \textbf{StackCap}~\cite{gu2017stack}, \textbf{Up-Down}~\cite{anderson2018bottom}, \textbf{LSTM-A}~\cite{yao2017boosting}, \textbf{GCN-LSTM}~\cite{yao2018exploring}, and \textbf{CAVP}~\cite{liu2018context}. Among these methods, SCST and Up-Down are two baselines where the more advanced self-critic reward and visual features are used. Compared with SCST, StackCap proposes a more complex RL-based reward for learning captions with more details. All of LSTM-A, GCN-LSTM, and CAVP try to exploit information of visual scene graphs, \eg, LSTM-A and GCN-LSTM exploit attributes and relationships information respectively, while CAVP tries to learn pairwise relationships in the decoder. Noteworthy, in GCN-LSTM, they set the batch size as $1,024$ and the training epoch as $250$, which is quite large compared with some other methods like Up-Down or CAVP, and is beyond our computation resources. For fair comparison, we also re-implemented a version of their work (since they do not publish the code), and set the batch size and training epoch both as $100$, such result is denoted as GCN-LSTM$^\dagger$ in Table~\ref{table:tab_kap}. In addition, the best result reported by GCN-LSTM is obtained by fusing two probabilities computed from two different kinds of relationships, which is denoted as GCN-LSTM$_{fuse}$, and our counterpart is denoted as SGAE$_{fuse}$.

\noindent\textbf{Analysis.}
From Table~\ref{table:tab_kap}, we can see that our single model achieves a new state-of-the-art score among all the compared methods in terms of CIDEr-D, which is $127.8$. And compared with GCN-LSTM$_{fuse}$, our fusion model SGAE$_{fuse}$ also achieves better performances. By exploiting the inductive bias in $\mathcal{D}$, even when our decoder or RL-reward is not as sophisticated as CVAP or StackCap, our method still has better performances. Moreover, our small batch size and fewer training epochs still lead to higher performances than GCN-LSTM, whose batch size and training epochs are much larger. Table~\ref{table:tab_online} reports the performances of different methods test on the official server. Compared with the published captioning methods (by the date of 16/11/2018), our single model has competitive performances and can achieve the highest CIDEr-D score.

\section{Conclusions}
We proposed to incorporate the language inductive bias --- a prior for more human-like language generation --- into the prevailing encoder-decoder framework for image captioning. In particular, we presented a novel unsupervised learning method: Scene Graph Auto-Encoder (SGAE), for embedding the inductive bias  into a dictionary, which can be shared as a re-encoder for language generation and significantly improve the performance of the encoder-decoder. We validate the SGAE-based framework by extensive ablations and comparisons with state-of-the-art performances on MS-COCO. As we believe that SGAE is a general solution for capturing the language inductive bias,  we are going to apply it in other vision-language tasks.

{\small
\bibliographystyle{ieee}
\bibliography{egbib}

\begin{thebibliography}{10}\itemsep=-1pt

\bibitem{anderson2016spice}
P.~Anderson, B.~Fernando, M.~Johnson, and S.~Gould.
\newblock Spice: Semantic propositional image caption evaluation.
\newblock In {\em European Conference on Computer Vision}, pages 382--398.
  Springer, 2016.

\bibitem{anderson2018bottom}
P.~Anderson, X.~He, C.~Buehler, D.~Teney, M.~Johnson, S.~Gould, and L.~Zhang.
\newblock Bottom-up and top-down attention for image captioning and visual
  question answering.
\newblock In {\em CVPR}, volume~3, page~6, 2018.

\bibitem{bahdanau2014neural}
D.~Bahdanau, K.~Cho, and Y.~Bengio.
\newblock Neural machine translation by jointly learning to align and
  translate.
\newblock {\em ICLR}, 2015.

\bibitem{banerjee2005meteor}
S.~Banerjee and A.~Lavie.
\newblock Meteor: An automatic metric for mt evaluation with improved
  correlation with human judgments.
\newblock In {\em Proceedings of the acl workshop on intrinsic and extrinsic
  evaluation measures for machine translation and/or summarization}, pages
  65--72, 2005.

\bibitem{battaglia2018relational}
P.~W. Battaglia, J.~B. Hamrick, V.~Bapst, A.~Sanchez-Gonzalez, V.~Zambaldi,
  M.~Malinowski, A.~Tacchetti, D.~Raposo, A.~Santoro, R.~Faulkner, et~al.
\newblock Relational inductive biases, deep learning, and graph networks.
\newblock {\em arXiv preprint arXiv:1806.01261}, 2018.

\bibitem{devlin2018bert}
J.~Devlin, M.-W. Chang, K.~Lee, and K.~Toutanova.
\newblock Bert: Pre-training of deep bidirectional transformers for language
  understanding.
\newblock {\em arXiv preprint arXiv:1810.04805}, 2018.

\bibitem{fang2015captions}
H.~Fang, S.~Gupta, F.~Iandola, R.~K. Srivastava, L.~Deng, P.~Doll{\'a}r,
  J.~Gao, X.~He, M.~Mitchell, J.~C. Platt, et~al.
\newblock From captions to visual concepts and back.
\newblock In {\em CVPR}, 2015.

\bibitem{gu2017stack}
J.~Gu, J.~Cai, G.~Wang, and T.~Chen.
\newblock Stack-captioning: Coarse-to-fine learning for image captioning.
\newblock {\em AAAI}, 2017.

\bibitem{he2016deep}
K.~He, X.~Zhang, S.~Ren, and J.~Sun.
\newblock Deep residual learning for image recognition.
\newblock In {\em Proceedings of the IEEE conference on computer vision and
  pattern recognition}, pages 770--778, 2016.

\bibitem{Hu_2018_CVPR}
R.~Hu, P.~Dollár, K.~He, T.~Darrell, and R.~Girshick.
\newblock Learning to segment every thing.
\newblock In {\em The IEEE Conference on Computer Vision and Pattern
  Recognition (CVPR)}, June 2018.

\bibitem{johnson2018image}
J.~Johnson, A.~Gupta, and L.~Fei-Fei.
\newblock Image generation from scene graphs.
\newblock {\em arXiv preprint}, 2018.

\bibitem{johnson2017clevr}
J.~Johnson, B.~Hariharan, L.~van~der Maaten, L.~Fei-Fei, C.~L. Zitnick, and
  R.~Girshick.
\newblock Clevr: A diagnostic dataset for compositional language and elementary
  visual reasoning.
\newblock In {\em Computer Vision and Pattern Recognition (CVPR), 2017 IEEE
  Conference on}, pages 1988--1997. IEEE, 2017.

\bibitem{johnson2015image}
J.~Johnson, R.~Krishna, M.~Stark, L.-J. Li, D.~Shamma, M.~Bernstein, and
  L.~Fei-Fei.
\newblock Image retrieval using scene graphs.
\newblock In {\em Proceedings of the IEEE conference on computer vision and
  pattern recognition}, pages 3668--3678, 2015.

\bibitem{karpathy2015deep}
A.~Karpathy and L.~Fei-Fei.
\newblock Deep visual-semantic alignments for generating image descriptions.
\newblock In {\em Proceedings of the IEEE conference on computer vision and
  pattern recognition}, pages 3128--3137, 2015.

\bibitem{kingma2014adam}
D.~P. Kingma and J.~Ba.
\newblock Adam: A method for stochastic optimization.
\newblock {\em arXiv preprint arXiv:1412.6980}, 2014.

\bibitem{kirillov2018panoptic}
A.~Kirillov, K.~He, R.~Girshick, C.~Rother, and P.~Doll{\'a}r.
\newblock Panoptic segmentation.
\newblock {\em arXiv preprint arXiv:1801.00868}, 2018.

\bibitem{klein2003accurate}
D.~Klein and C.~D. Manning.
\newblock Accurate unlexicalized parsing.
\newblock In {\em Proceedings of the 41st Annual Meeting on Association for
  Computational Linguistics-Volume 1}, pages 423--430. Association for
  Computational Linguistics, 2003.

\bibitem{krishna2017visual}
R.~Krishna, Y.~Zhu, O.~Groth, J.~Johnson, K.~Hata, J.~Kravitz, S.~Chen,
  Y.~Kalantidis, L.-J. Li, D.~A. Shamma, et~al.
\newblock Visual genome: Connecting language and vision using crowdsourced
  dense image annotations.
\newblock {\em International Journal of Computer Vision}, 123(1):32--73, 2017.

\bibitem{kulkarni2013babytalk}
G.~Kulkarni, V.~Premraj, V.~Ordonez, S.~Dhar, S.~Li, Y.~Choi, A.~C. Berg, and
  T.~L. Berg.
\newblock Babytalk: Understanding and generating simple image descriptions.
\newblock In {\em CVPR}, 2011.

\bibitem{kuznetsova2012collective}
P.~Kuznetsova, V.~Ordonez, A.~C. Berg, T.~L. Berg, and Y.~Choi.
\newblock Collective generation of natural image descriptions.
\newblock In {\em Proceedings of the 50th Annual Meeting of the Association for
  Computational Linguistics: Long Papers-Volume 1}, pages 359--368. Association
  for Computational Linguistics, 2012.

\bibitem{lake2017building}
B.~M. Lake, T.~D. Ullman, J.~B. Tenenbaum, and S.~J. Gershman.
\newblock Building machines that learn and think like people.
\newblock {\em Behavioral and Brain Sciences}, 40, 2017.

\bibitem{li2011composing}
S.~Li, G.~Kulkarni, T.~L. Berg, A.~C. Berg, and Y.~Choi.
\newblock Composing simple image descriptions using web-scale n-grams.
\newblock In {\em Proceedings of the Fifteenth Conference on Computational
  Natural Language Learning}, pages 220--228. Association for Computational
  Linguistics, 2011.

\bibitem{li2015gated}
Y.~Li, D.~Tarlow, M.~Brockschmidt, and R.~Zemel.
\newblock Gated graph sequence neural networks.
\newblock {\em arXiv preprint arXiv:1511.05493}, 2015.

\bibitem{lin2004rouge}
C.-Y. Lin.
\newblock Rouge: A package for automatic evaluation of summaries.
\newblock {\em Text Summarization Branches Out}, 2004.

\bibitem{lin2014microsoft}
T.-Y. Lin, M.~Maire, S.~Belongie, J.~Hays, P.~Perona, D.~Ramanan,
  P.~Doll{\'a}r, and C.~L. Zitnick.
\newblock Microsoft coco: Common objects in context.
\newblock In {\em European conference on computer vision}, pages 740--755.
  Springer, 2014.

\bibitem{liu2018context}
D.~Liu, Z.-J. Zha, H.~Zhang, Y.~Zhang, and F.~Wu.
\newblock Context-aware visual policy network for sequence-level image
  captioning.
\newblock In {\em 2018 ACM Multimedia Conference on Multimedia Conference},
  pages 1416--1424. ACM, 2018.

\bibitem{lu2017knowing}
J.~Lu, C.~Xiong, D.~Parikh, and R.~Socher.
\newblock Knowing when to look: Adaptive attention via a visual sentinel for
  image captioning.
\newblock In {\em Proceedings of the IEEE Conference on Computer Vision and
  Pattern Recognition (CVPR)}, volume~6, page~2, 2017.

\bibitem{lu2018neural}
J.~Lu, J.~Yang, D.~Batra, and D.~Parikh.
\newblock Neural baby talk.
\newblock In {\em Proceedings of the IEEE Conference on Computer Vision and
  Pattern Recognition}, pages 7219--7228, 2018.

\bibitem{marcheggiani2017encoding}
D.~Marcheggiani and I.~Titov.
\newblock Encoding sentences with graph convolutional networks for semantic
  role labeling.
\newblock In {\em Proceedings of the 2017 Conference on Empirical Methods in
  Natural Language Processing}, pages 1506--1515, 2017.

\bibitem{marr1982vision}
D.~Marr.
\newblock Vision: A computational investigation into the human representation
  and processing of visual information. mit press.
\newblock {\em Cambridge, Massachusetts}, 1982.

\bibitem{mitchell2012midge}
M.~Mitchell, X.~Han, J.~Dodge, A.~Mensch, A.~Goyal, A.~Berg, K.~Yamaguchi,
  T.~Berg, K.~Stratos, and H.~Daum{\'e}~III.
\newblock Midge: Generating image descriptions from computer vision detections.
\newblock In {\em Proceedings of the 13th Conference of the European Chapter of
  the Association for Computational Linguistics}, pages 747--756. Association
  for Computational Linguistics, 2012.

\bibitem{papineni2002bleu}
K.~Papineni, S.~Roukos, T.~Ward, and W.-J. Zhu.
\newblock Bleu: a method for automatic evaluation of machine translation.
\newblock In {\em Proceedings of the 40th annual meeting on association for
  computational linguistics}, pages 311--318. Association for Computational
  Linguistics, 2002.

\bibitem{ranzato2015sequence}
M.~Ranzato, S.~Chopra, M.~Auli, and W.~Zaremba.
\newblock Sequence level training with recurrent neural networks.
\newblock 2015.

\bibitem{redmon2017yolo9000}
J.~Redmon and A.~Farhadi.
\newblock Yolo9000: Better, faster, stronger.
\newblock In {\em 2017 IEEE Conference on Computer Vision and Pattern
  Recognition (CVPR)}, pages 6517--6525. IEEE, 2017.

\bibitem{ren2015faster}
S.~Ren, K.~He, R.~Girshick, and J.~Sun.
\newblock Faster r-cnn: Towards real-time object detection with region proposal
  networks.
\newblock In {\em Advances in neural information processing systems}, pages
  91--99, 2015.

\bibitem{rennie2017self}
S.~J. Rennie, E.~Marcheret, Y.~Mroueh, J.~Ross, and V.~Goel.
\newblock Self-critical sequence training for image captioning.
\newblock In {\em CVPR}, volume~1, page~3, 2017.

\bibitem{schuster2015generating}
S.~Schuster, R.~Krishna, A.~Chang, L.~Fei-Fei, and C.~D. Manning.
\newblock Generating semantically precise scene graphs from textual
  descriptions for improved image retrieval.
\newblock In {\em Proceedings of the fourth workshop on vision and language},
  pages 70--80, 2015.

\bibitem{sukhbaatar2015end}
S.~Sukhbaatar, J.~Weston, R.~Fergus, et~al.
\newblock End-to-end memory networks.
\newblock In {\em Advances in neural information processing systems}, pages
  2440--2448, 2015.

\bibitem{teney2017graph}
D.~Teney, L.~Liu, and A.~van~den Hengel.
\newblock Graph-structured representations for visual question answering.
\newblock In {\em 2017 IEEE Conference on Computer Vision and Pattern
  Recognition (CVPR)}, pages 3233--3241. IEEE, 2017.

\bibitem{vedantam2015cider}
R.~Vedantam, C.~Lawrence~Zitnick, and D.~Parikh.
\newblock Cider: Consensus-based image description evaluation.
\newblock In {\em Proceedings of the IEEE conference on computer vision and
  pattern recognition}, pages 4566--4575, 2015.

\bibitem{vinyals2016matching}
O.~Vinyals, C.~Blundell, T.~Lillicrap, D.~Wierstra, et~al.
\newblock Matching networks for one shot learning.
\newblock In {\em Advances in Neural Information Processing Systems}, pages
  3630--3638, 2016.

\bibitem{vinyals2015show}
O.~Vinyals, A.~Toshev, S.~Bengio, and D.~Erhan.
\newblock Show and tell: A neural image caption generator.
\newblock In {\em CVPR}, 2015.

\bibitem{N18-1037}
Y.-S. Wang, C.~Liu, X.~Zeng, and A.~Yuille.
\newblock Scene graph parsing as dependency parsing.
\newblock In {\em Proceedings of the 2018 Conference of the North American
  Chapter of the Association for Computational Linguistics: Human Language
  Technologies, Volume 1 (Long Papers)}, pages 397--407. Association for
  Computational Linguistics, 2018.

\bibitem{xiong2016dynamic}
C.~Xiong, S.~Merity, and R.~Socher.
\newblock Dynamic memory networks for visual and textual question answering.
\newblock In {\em International conference on machine learning}, pages
  2397--2406, 2016.

\bibitem{xu2017scene}
D.~Xu, Y.~Zhu, C.~B. Choy, and L.~Fei-Fei.
\newblock Scene graph generation by iterative message passing.
\newblock In {\em Proceedings of the IEEE Conference on Computer Vision and
  Pattern Recognition}, volume~2, 2017.

\bibitem{xu2015show}
K.~Xu, J.~Ba, R.~Kiros, K.~Cho, A.~Courville, R.~Salakhudinov, R.~Zemel, and
  Y.~Bengio.
\newblock Show, attend and tell: Neural image caption generation with visual
  attention.
\newblock In {\em International conference on machine learning}, pages
  2048--2057, 2015.

\bibitem{yang2018graph}
J.~Yang, J.~Lu, S.~Lee, D.~Batra, and D.~Parikh.
\newblock Graph r-cnn for scene graph generation.
\newblock In {\em European Conference on Computer Vision}, pages 690--706.
  Springer, 2018.

\bibitem{yang2018shuffle}
X.~Yang, H.~Zhang, and J.~Cai.
\newblock Shuffle-then-assemble: Learning object-agnostic visual relationship
  features.
\newblock In {\em European Conference on Computer Vision}, pages 38--54.
  Springer, 2018.

\bibitem{yao2018exploring}
T.~Yao, Y.~Pan, Y.~Li, and T.~Mei.
\newblock Exploring visual relationship for image captioning.
\newblock In {\em Computer Vision--ECCV 2018}, pages 711--727. Springer, 2018.

\bibitem{yao2017boosting}
T.~Yao, Y.~Pan, Y.~Li, Z.~Qiu, and T.~Mei.
\newblock Boosting image captioning with attributes.
\newblock In {\em IEEE International Conference on Computer Vision, ICCV},
  pages 22--29, 2017.

\bibitem{yosinski2014transferable}
J.~Yosinski, J.~Clune, Y.~Bengio, and H.~Lipson.
\newblock How transferable are features in deep neural networks?
\newblock In {\em Advances in neural information processing systems}, pages
  3320--3328, 2014.

\bibitem{zellers2018neural}
R.~Zellers, M.~Yatskar, S.~Thomson, and Y.~Choi.
\newblock Neural motifs: Scene graph parsing with global context.
\newblock In {\em Proceedings of the IEEE Conference on Computer Vision and
  Pattern Recognition}, pages 5831--5840, 2018.

\bibitem{zhang2017visual}
H.~Zhang, Z.~Kyaw, S.-F. Chang, and T.-S. Chua.
\newblock Visual translation embedding network for visual relation detection.
\newblock In {\em CVPR}, volume~1, page~5, 2017.

\bibitem{zhang2018learning}
Y.~Zhang, J.~Hare, and A.~Pr{\"u}gel-Bennett.
\newblock Learning to count objects in natural images for visual question
  answering.
\newblock In {\em ICLR}, 2018.

\end{thebibliography}
}

\onecolumn
This supplementary document will further detail the following aspects in the main paper: A. Network Architecture, B. Details of Scene Graphs, C. More Qualitative Examples.

\section{Network Architecture}
Here, we introduce the detailed network architectures of all the components in our model, which includes Graph Convolutional Network (GCN), Multi-modal Graph Convolutional Network (MGCN), Dictionary, and Decoders.
\subsection{Graph Convolutional Network}

\begin{table*}[t]
\begin{center}
\caption{The details of GCN.}
\label{table:tab_gcn}
\begin{tabular}{|c|c|c|c|c|c|}
		\hline
		   \textbf{Index}&\textbf{Input}&\textbf{Operation}&\textbf{Output}&\textbf{Trainable Parameters}\\ \hline
		   (1)  &    -    &  object label  & $l_o$ (10,102) & - \\ \hline
		   (2)  &    -    &  relation label  & $l_r$ (10,102) & - \\ \hline
		   (3)  &    -    &  attribute label  & $l_a$ (10,102) & - \\ \hline
		   (4)  &   (1)    &  word embedding $\bm{W}_{\Sigma_S} l_o$   & $\bm{e}_o$ (1,000) & $\bm{W}_{\Sigma_S}$ (1,000 $\times$ 10,102) \\ \hline
		   (5)  &   (2)    &  word embedding $\bm{W}_{\Sigma_S} l_r$   & $\bm{e}_r$ (1,000) & $\bm{W}_{\Sigma_S}$ (1,000 $\times$ 10,102)\\ \hline
		   (6)  &   (3)   &  word embedding $\bm{W}_{\Sigma_S} l_a$   & $\bm{e}_a$ (1,000) & $\bm{W}_{\Sigma_S}$ (1,000 $\times$ 10,102)  \\ \hline
		   (7)  & (4),(5) &  relationship embedding (Eq.(6)) & $\bm{x}_r$ (1,000) & $g_r$ (3,000 $\rightarrow$ 1,000)   \\ \hline
		   (8)  & (4),(6) &  attribute embedding (Eq.(7)) & $\bm{x}_a$ (1,000)   & $g_a$ (2,000 $\rightarrow$ 1,000)\\ \hline
		   (9)  & (4),(5) &  object embedding (Eq.(8)) & $\bm{x}_o$ (1,000)& $g_s$,$g_o$ (3,000 $\rightarrow$ 1,000) \\ \hline
\end{tabular}
\end{center}
\end{table*}

In Section~4.2 of the main paper, we show how to use GCN to compute three embeddings by given a sentence scene graph, and the operations of this GCN are listed in Table~\ref{table:tab_gcn}. In Table~\ref{table:tab_gcn} (1) to (3), the object label $l_o$, relation label $l_r$, and attribute label $l_a$ are all one-hot vectors. And the word embedding matrix $\bm{W}_{\Sigma_S}\in \mathbb{R}^{1,000\times10,102}$ is used to map these one-hot vectors into continuous vector representations in Table~\ref{table:tab_gcn} (4) to (6). The second dimension of $\bm{W}_{\Sigma_S}$ is the total number of object, relation, and attribute categories among all the sentence scene graphs. For $g_r$,  $g_a$, $g_o$, and $g_s$ in Table~\ref{table:tab_gcn} (7) to (9), all of them own the same structure with independent parameters: a fully-connected layer, followed by an ReLU. The notation $g_r$ ($D_{in}$ $\rightarrow$ $D_{out}$) denote that the input dimension is $D_{in}$, and output dimension is $D_{out}$.

\subsection{Multi-modal Graph Convolutional Network}
\begin{table*}[t]
\begin{center}
\caption{The details of MGCN}
\label{table:tab_mgcn}
\begin{tabular}{|c|c|c|c|c|c|}
		\hline
		   \textbf{Index}&\textbf{Input}&\textbf{Operation}&\textbf{Output}&\textbf{Trainable Parameters}\\ \hline
		   (1)  &    -   &    object RoI feature  & $\bm{v}_o$ (2,048) & -  \\ \hline
		   (2)  &    -   &    relation RoI feature  & $\bm{v}_r$ (2,048) & -  \\ \hline
		   (3)  &    -    &  object label  & $l_o$ (472) & - \\ \hline
		   (4)  &    -    &  relation label  & $l_r$ (472) & - \\ \hline
		   (5)  &    -    &  attribute label  & $l_a$ (472) & - \\ \hline
		   (6)  &   (3)    &  word embedding $\bm{W}_{\Sigma_I} l_o$   & $\bm{e}_o$ (1,000) & $\bm{W}_{\Sigma_I}$ (1,000 $\times$ 472)  \\ \hline
		   (7)  &   (4)    &  word embedding $\bm{W}_{\Sigma_I} l_r$   & $\bm{e}_r$ (1,000) & $\bm{W}_{\Sigma_I}$ (1,000 $\times$ 472)  \\ \hline
		   (8)  &   (5)    &  word embedding $\bm{W}_{\Sigma_I} l_a$   & $\bm{e}_a$ (1,000) & $\bm{W}_{\Sigma_I}$ (1,000 $\times$ 472)  \\ \hline
		   (9)  &(1),(6)  &  
		   \begin{tabular}{c}
		         feature fusion \\
		         $\text{ReLU}(\bm{W}_1^o \bm{e}_{o} + \bm{W}_2^o \bm{v}_{o})$ \\
		         $- (\bm{W}_1^o \bm{e}_{o} - \bm{W}_2^o \bm{v}_{o})^2$
		   \end{tabular}
		   & $\bm{u}_o$ (1,000) &
		   \begin{tabular}{c}
		         $\bm{W}_1^o$ (1,000 $\times$ 1,000) \\
		         $\bm{W}_2^o$ (1,000 $\times$ 2,048)
		   \end{tabular}
		    \\ \hline
		   (10)  &(2),(7)  &  
		   \begin{tabular}{c}
		         feature fusion \\
		         $\text{ReLU}(\bm{W}_1^r \bm{e}_{r} + \bm{W}_2^r \bm{v}_{r})$ \\
		         $- (\bm{W}_1^r \bm{e}_{r} - \bm{W}_2^r \bm{v}_{r})^2$
		   \end{tabular}
		   & $\bm{u}_r$ (1,000) &
		   \begin{tabular}{c}
		         $\bm{W}_1^r$ (1,000 $\times$ 1,000) \\
		         $\bm{W}_2^r$ (1,000 $\times$ 2,048)
		   \end{tabular}  \\ \hline
		   (11)  &(1),(8)  &  
		   \begin{tabular}{c}
		         feature fusion \\
		         $\text{ReLU}(\bm{W}_1^a \bm{e}_{a} + \bm{W}_2^a \bm{v}_{o})$ \\
		         $ - (\bm{W}_1^a \bm{e}_{a} - \bm{W}_2^a \bm{v}_{o})^2$
		   \end{tabular}
		   & $\bm{u}_a$ (1,000) &
		   \begin{tabular}{c}
		         $\bm{W}_1^a$ (1,000 $\times$ 1,000) \\
		         $\bm{W}_2^a$ (1,000 $\times$ 2,048)
		   \end{tabular}
		   \\ \hline
		   (12)  &(9),(10)  &  relationship embedding (Eq.~\ref{equ:equ_mgcnr}) & $\bm{v}_r^{'}$ (1,000) & $f_r$ (3,000 $\rightarrow$ 1,000)  \\ \hline
		   (13)  &(9),(11)  &  attribute embedding (Eq.~\ref{equ:equ_mgcna}) & $\bm{v}_a^{'}$ (1,000)  & $f_a$ (2,000 $\rightarrow$ 1,000) \\ \hline
		  (14)  &(9),(10)  &  object embedding (Eq.~\ref{equ:equ_mgcno}) & $\bm{v}_o^{'}$ (1,000) & $f_s$,$f_o$ (3,000 $\rightarrow$ 1,000)  \\ \hline
\end{tabular}
\end{center}
\end{table*}
In Section~5.1 of the main paper, we briefly discuss the MGCN, and here we list its details in Table~\ref{table:tab_mgcn}. Besides the labels of objects, relations, and attributes, the input of MGCN also include object and relation RoI features, as shown in Table~\ref{table:tab_mgcn} (1) to (5). The RoI features are extracted from a pre-trained Faster Rcnn~\cite{ren2015faster}, $v_r$ is the feature pooled from a region which cover the `subject' and `object'. The word embedding matrix used here in Table~\ref{table:tab_mgcn} (6) to (8) is $\bm{W}_{\Sigma_I} \in \mathbb{R}^{1,000\times472}$, which is different from the one used in GCN. In Table~\ref{table:tab_mgcn} (9) to (11), feature fusion proposed by~\cite{zhang2018learning} is implemented for fusing word embedding and visual feature together. 
Compared with Eq.~(6) to Eq.~(8) in the main paper, MGCN has the following modifications for computing relationship, attribute, and object embeddings: word embeddings $\bm{e}$ are substituted by fused embeddings $\bm{u}$; and $g$ is substituted by $f$, which is also a function of a fully-connected layer, followed by an ReLU. With these modifications, we can formulate the computations of three embeddings in MGCN as:

\noindent\textbf{Relationship Embedding $\bm{v}_{r_{ij}}^{'}$} (Table~\ref{table:tab_mgcn} (12)):
\begin{equation}
\label{equ:equ_mgcnr}
    \bm{v}_{r_{ij}}^{'}=f_r(\bm{u}_{o_i}, \bm{u}_{r_{ij}}, \bm{u}_{o_j}).
\end{equation}

\noindent\textbf{Attribute Embedding $\bm{v}_{a_i}^{'}$}(Table~\ref{table:tab_mgcn} (13)):
\begin{equation}
   \bm{v}_{a_i}^{'} =\frac{1}{Na_i}\sum_{l=1}^{Na_i}f_a(\bm{u}_{o_i},\bm{u}_{a_{i,l}}).
\label{equ:equ_mgcna}
\end{equation}

\noindent\textbf{Object Embedding $\bm{v}_{o_i}^{'}$}(Table~\ref{table:tab_mgcn} (14)):
\begin{equation}
\label{equ:equ_mgcno}
    \bm{v}_{o_i}^{'} = \frac{1}{Nr_i}[\sum_{o_j\in sbj(o_i)}f_s(\bm{u}_{o_i},\bm{u}_{o_j},\bm{u}_{r_{ij}}) +\sum_{o_k\in obj(o_i)}f_o(\bm{u}_{o_k},\bm{u}_{o_i},\bm{u}_{r_{ki}})].
\end{equation}

\subsection{Dictionary}
The re-encoder function in Section~4.3 is used to re-encode a new representation $\hat{\bm{x}}$ from an index vector $\bm{x}$ and a dictionary $\mathcal{D}$, such operation is given in Table~\ref{table:tab_dict}. As shown in Table~\ref{table:tab_dict} (2) and (3) respectively, by given an index vector $\bm{x}$, we first do inner produce between each element in $\bm{D}$ with $\bm{x}$ and then use softmax to normalize the computed results. At last, the re-encoded $\hat{\bm{x}}$ is the weighted sum of each atom in $\bm{D}$ as $\sum_{k=1}^K\alpha_k\bm{d}_k$, $K$ is set as 10,000.
\begin{table*}[t]
\begin{center}
\caption{The details of the re-encoder function.}
\label{table:tab_dict}
\begin{tabular}{|c|c|c|c|c|}
		\hline
		   \textbf{Index}&\textbf{Input}&\textbf{Operation}&\textbf{Output}&\textbf{Trainable Parameters}\\ \hline
		   (1)  & index vector &  -  & $\bm{x}$ (1,000) & - \\ \hline
		   (2)  & (1)     & inner product $\bm{D}^T\bm{x}$ & $\bm{\alpha}$ (10,000) & $\bm{D}$(1,000 $\times$ 10,000) \\ \hline
		   (3)  & (2)          & softmax & $\bm{\alpha}$ (10,000) & - \\ \hline
		   (4)  & (3)          & weighted sum $\bm{D}\bm{\alpha}$ & $\hat{\bm{x}}$(1,000) & $\bm{D}$(1,000 $\times$ 10,000)\\ \hline
\end{tabular}
\end{center}
\end{table*}

\subsection{Decoders}
We followed the language decoder proposed by~\cite{anderson2018bottom} to set our two decoders of Eq.~(4) and Eq.~(5) in the main paper. Both decoders have the same architecture, as shown in Table~\ref{table:tab_dec}, except for the different embedding sets used as their inputs. For convenience, we introduce the decoders' common architecture without differentiating them between Eq.~(4) and Eq.~(5), and then detail the difference between them at the end of this section.

The implemented decoder contains two LSTM layers and one attention module. The input of the first LSTM contains the concatenation of three terms: word embedding vector $\bm{W}_{\Sigma}\bm{w}_{t-1}$, mean pooling of embedding set $\bar{\bm{z}}$, and the output of the second LSTM $\bm{h}_{t-1}^2$. We use them as input since they can provide abundant accumulated context information. Then, an index vector $\bm{h}_{t-1}^1$ is created by LSTM$_1$ in Table~\ref{table:tab_dec} (7), which will be used to instruct the decoder to put attention on suitable embedding of $\mathcal{Z}$ by an attention module. Given $\mathcal{Z}$ and $\bm{h}_{t-1}^1$, the formulations in Table~\ref{table:tab_dec} (8) and (9) can be applied for computing a $M$-dimension attention distribution $\bm{\beta}$, and then we can create the attended embedding $\hat{\bm{z}}$ by weighted sum as in (10). By inputting $\hat{\bm{z}}$ and $\bm{h}_{t-1}^1$ into LSTM$_2$ and implementing (11) to (13), the word distribution $P_t$ can be got for sampling a word at time $t$.

For two decoders in Eq.~(4) and Eq.~(5), they only differ in using different embedding sets $\mathcal{Z}$ as inputs. In SGAE (Eq.~(5)), $\mathcal{Z}$ is set as $\hat{\mathcal{X}}$. While in SGAE-based encoder-decoder (Eq.~(4)), we have a small modification that the vector $\bm{z} \in \mathcal{Z}$ is set as follows: $\bm{z}=[\bm{v}',\hat{\bm{v}}]$, where $\bm{v}' \in \mathcal{V}'$ ($\mathcal{V}'$ is the scene graph-modulated feature set in Section~5.1), and $\hat{\bm{v}} \in \hat{\mathcal{V}}$ ($\hat{\mathcal{V}}$ is the re-encoded feature set in Section~5.1).

\begin{table*}[t]
\begin{center}
\caption{The details of the common structure of the two decoders.}
\label{table:tab_dec}
\begin{tabular}{|c|c|c|c|c|}
		\hline
		   \textbf{Index}&\textbf{Input}&\textbf{Operation}&\textbf{Output}&\textbf{Trainable Parameters}\\ \hline
		   (1)  & - &  word label     &  $\bm{w}_{t-1}$ (10,369) & - \\ \hline
		   (2)  & -  &  embedding set  & $\mathcal{Z}$ (1,000 $\times$ M) & -  \\ \hline
		   (3)  & - & output of LSTM$_2$ & $\bm{h}_{t-1}^2$ (1,000) & -  \\ \hline
		   (4)  & (1)   &  word embedding  $\bm{W}_{\Sigma}\bm{w}_{t-1}$  &  $\bm{e}_{t-1}$ (1,000) & $\bm{W}_{\Sigma}$ (1,000 $\times$ 10,369) \\ \hline
		   (5)  & (2)    & mean pooling & $\bar{\bm{z}}$ (1,000) & - \\ \hline 
		   (6)  & (3),(4),(5) & concatenate & $\bm{i}_t$ (3,000) & - \\ \hline 
		   (7)  & (6)  & LSTM$_1$ $(\bm{i}_t;\bm{h}_{t-1}^1)$  & $\bm{h}_{t}^1$ (1,000) &   LSTM$_1$ (3,000 $\rightarrow$ 1,000)\\ \hline 
		   (8)  & (2),(7) &  $\bm{w}_a\tanh(\bm{W}_z\bm{z}_{m}+\bm{W}_h\bm{h}_{t}^1)$ & $\bm{\beta}$ (M) &
		   \begin{tabular}{c}
		         $\bm{w}_a$ (512), $\bm{W}_z$ (512$\times$1,000) \\
		         $\bm{W}_h$(512$\times$1,000)
		   \end{tabular}
		   \\ \hline 
		   (9)  & (8)    & softmax & $\bm{\beta}$ (M) & - \\ \hline
		   (10)  & (9),(2) & weighted sum $\mathcal{Z}\bm{\beta}$ & $\hat{\bm{z}}$ (1,000) & - \\ \hline 
		   (11) & (7),(10) & LSTM$_2$ $([\bm{h}_t^1,\hat{\bm{z}}];\bm{h}_{t-1}^2)$ & $\bm{h}_t^2$ (1,000) & LSTM$_1$ (3,000 $\rightarrow$ 1,000) \\ \hline
		   (12) & (11)    &
		   $\bm{W}_{p}\bm{h}_{t}^2+\bm{b}_{p}$ & $\bm{p}_t$ (10,369) &
		   \begin{tabular}{c}
		         $\bm{W}_{p}$ (10,369 $\times$ 1,000) \\
		         $\bm{b}_{p}$ (10,369)
		   \end{tabular}
		   \\ \hline
		   (13) & (12) & softmax & $P_t$ (10,369) & - \\ \hline
		   
\end{tabular}
\end{center}
\end{table*}

\section{Details of Scene Graph}
\subsection{Sentence Scene Graph}
For each sentence, we directly implemented the software provided by~\cite{anderson2016spice} to parse its scene graph. And we filtered them by removing objects, relationships, and attributes which appear less than 10 among all the parsed scene graphs. After filtering, there are 5,364 objects, 1,308 relationships, and 3,430 attributes remaining. We grouped them together and used word embedding matrix $\bm{W}_{\Sigma_S}$ in Table~\ref{table:tab_gcn} to transform nodes' labels to continuous vector representations.

\subsection{Image Scene Graph}
\begin{table*}[t]
\begin{center}
\caption{The details of attribute classifier.}
\label{table:tab_attr}
\begin{tabular}{|c|c|c|c|c|}
		\hline
		   \textbf{Index}&\textbf{Input}&\textbf{Operation}&\textbf{Output}&\textbf{Trainable Parameters}\\ \hline
		   (1)  & object RoI feature &  -  & $\bm{v}$ (2,048) & - \\ \hline
		   (2)  & (1)     & fc & $\bm{f}_1$ (1,000) & fc(2,048 $\rightarrow$ 1,000) \\ \hline
		   (3)  & (2)     & ReLU & $\bm{f}_1$ (1,000) & - \\ \hline
		   (4)  & (3)          & fc & $\bm{f}_2$ (103) & fc(1,000 $\rightarrow$ 103) \\ \hline
		   (5)  & (4)          & softmax  & $P_a$ (103) & -\\ \hline
\end{tabular}
\end{center}
\end{table*}
Compared with sentence scene graphs, the parsing of image scene graphs is more complicated that we used Faster-RCNN as the object detector~\cite{ren2015faster} to detect and classify objects, MOTIFS relationship detector~\cite{zellers2018neural} to classify relationships between objects, and one simple attribute classifier to predict attributes. The details of them are given as follows.

\noindent\textbf{Object Detector:}
For detecting objects and extracting their RoI features, we followed~\cite{anderson2018bottom} to train Faster-RCNN. After training, we used 0.7 as the IoU threshold for proposal NMS, and 0.3 as threshold for object NMS. Also, we selected at least 10 objects and at most 100 objects for each image. RoI pooling was used to extract these objects' features, which will be used as the input to the relationship classifier, attribute classifier, and MGCN.

\noindent\textbf{Relationship Classifier:}
We used the LSTM structure proposed in~\cite{zellers2018neural} as our relationship classifier. After training, we predicted a relationship for each two objects whose IoU is larger than 0.2.

\noindent\textbf{Attribute Classifier:}
The detail structure of our attribute classifier is given in Table~\ref{table:tab_attr}. After training, we predicted top-3 attributes for each object.

For each image, by using predicted objects, relationships and attributes, an image scene graph can be built. As detailed in Section~6.1 of the main paper, the total number of used objects, relationships, and attributes here is 472, thus we used a 472 $\times$ 1,000 word embedding matrix to transform the nodes' labels into the continuous vectors as in Table~\ref{table:tab_mgcn} (6) to (8).

The codes and all these parsed scene graphs will be published for further research upon paper acceptance.

\section{More Qualitative Examples}
\begin{figure*}[t]
\centering
\includegraphics[width=0.9\linewidth,trim = 5mm 3mm 5mm 5mm,clip]{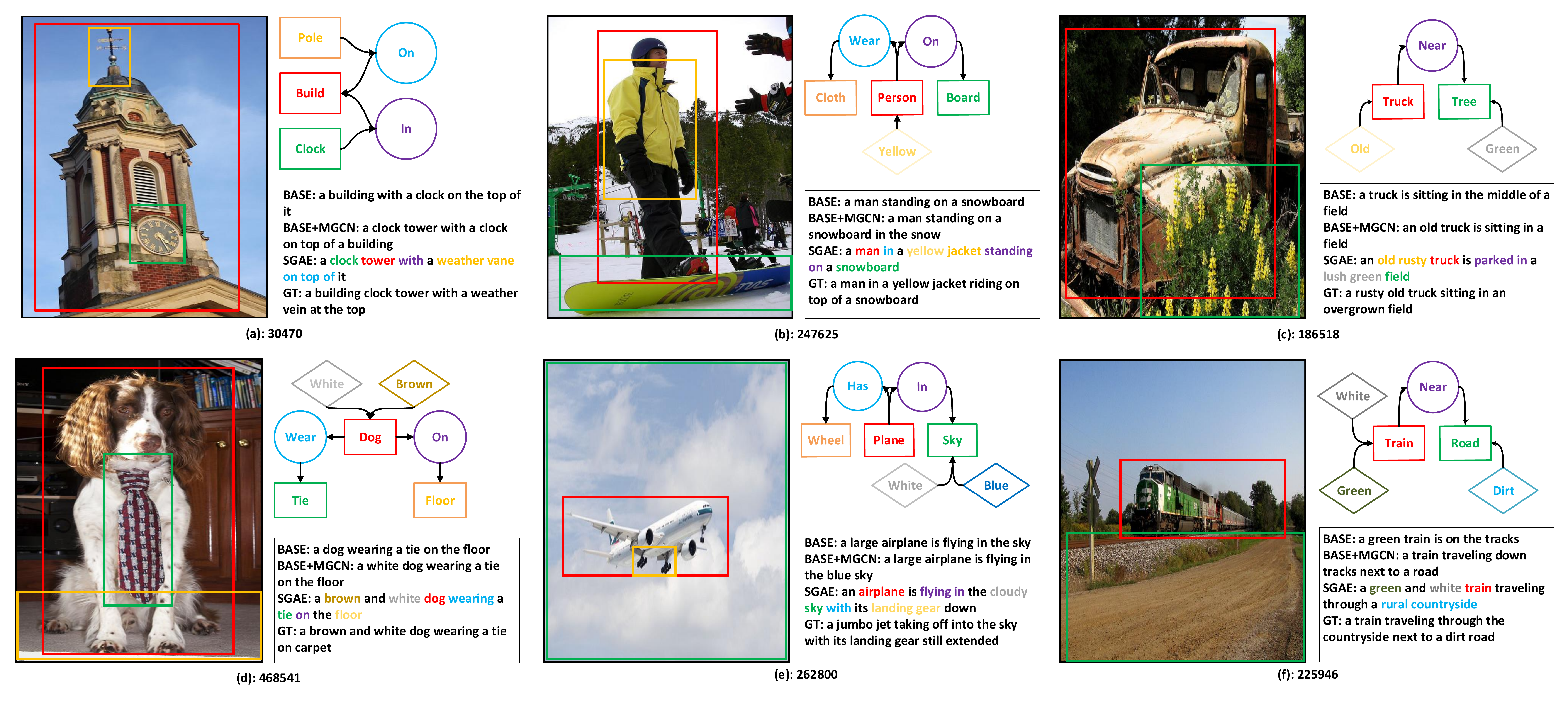}
   \caption{Qualitative examples of different baselines. For each figure, the image scene graph is pruned to avoid clutter.  The id refers to the image id in MS-COCO. Word colors correspond to nodes in the detected scene graphs. }
   \vspace{-0.1in}
\label{fig:fig_supp}
\end{figure*}

\begin{figure*}[t]
\centering
\includegraphics[width=0.85\linewidth,trim = 5mm 3mm 5mm 5mm,clip]{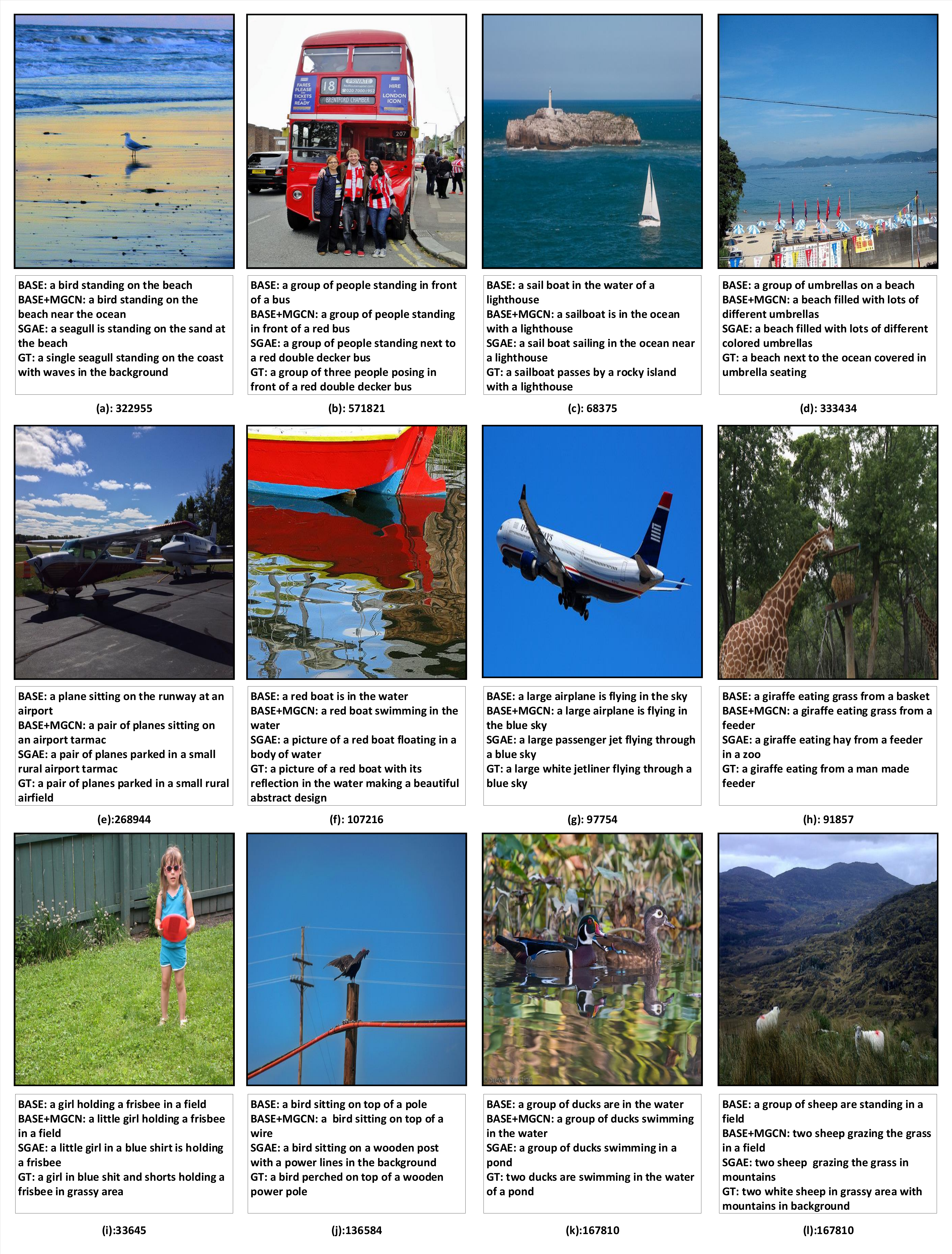}
   \caption{12 qualitative examples of different baselines.}
   \vspace{-0.1in}
\label{fig:fig_supp2}
\end{figure*}

Figure~\ref{fig:fig_supp} and~\ref{fig:fig_supp2} show more examples of generated captions of our methods and some baselines. We can find that the captions generated by SGAE prefer to use some more accurate words to describe the appeared objects, attributes, relationships or scenes. For instance, in Figure~\ref{fig:fig_supp} (a), the object `weather vane' is used while this object is not accurately recognized by the object detector; in Figure~\ref{fig:fig_supp} (c), SGAE prefers the attribute `old rusty'; in Figure~\ref{fig:fig_supp2} SGAE describes the relationship between boat with water as `floating' instead of `swimming'; and in Figure~\ref{fig:fig_supp2}, the scene `mountains' is inferred by using SGAE.

\end{document}